\documentclass[10pt,twocolumn,letterpaper]{article}
\usepackage[pagenumbers]{cvpr} 

\usepackage[dvipsnames]{xcolor}

\definecolor{cvprblue}{rgb}{0.21,0.49,0.74}
\usepackage[pagebackref,breaklinks,colorlinks,citecolor=cvprblue]{hyperref}
\usepackage{algorithm}
\usepackage{algorithmic}
\usepackage{listings}
\usepackage{graphicx}
\usepackage{multirow}
\usepackage{colortbl}
\usepackage{xcolor}
\usepackage{adjustbox}
\usepackage{tabularray}
\definecolor{tabfirst}{HTML}{E6E6E6} 
\definecolor{tabsecond}{rgb}{1, 0.85, 0.7} 
\definecolor{tabthird}{rgb}{1, 1, 0.7} 
\usepackage[percent]{overpic}

\definecolor{wo_text_mark}{HTML}{A13332}
\definecolor{raw_mark}{HTML}{3333A1} 
\definecolor{co_mark}{HTML}{32A2A3} 
\definecolor{all_mark}{HTML}{599A59} 
\definecolor{rand_co_mark}{HTML}{989934} 
\definecolor{rand_all_mark}{HTML}{B368B4} 

\begin{document}

\title{\begin{adjustbox}{valign=c}\includegraphics[width=1.2cm]{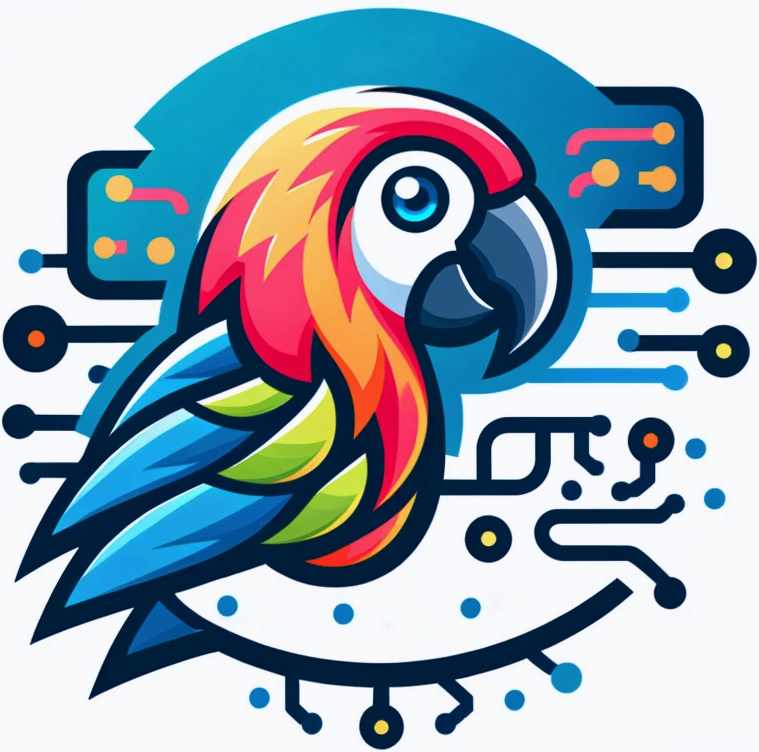}\end{adjustbox} Parrot Captions Teach CLIP to Spot Text}

\author{Yiqi Lin\textsuperscript{\mdseries1*} 
Conghui He\textsuperscript{1*\dag}  
Alex Jinpeng Wang\textsuperscript{\mdseries2*}  
Bin Wang\textsuperscript{\mdseries1*} \\
Weijia Li\textsuperscript{3}  
Mike Zheng Shou\textsuperscript{2}
 \\ \\
\textsuperscript{1} Shanghai AI Laboratory 
\textsuperscript{2} Show Lab, National University of Singapore \\
\textsuperscript{3} Sun Yat-Sen University \\ 
{\tt\small  \url{https://linyq17.github.io/CLIP-Parrot-Bias/}
}
}

\twocolumn[{
\renewcommand\twocolumn[1][]{#1}
\maketitle
\begin{center}
    \captionsetup{type=figure}
    \includegraphics[width=0.98\linewidth]{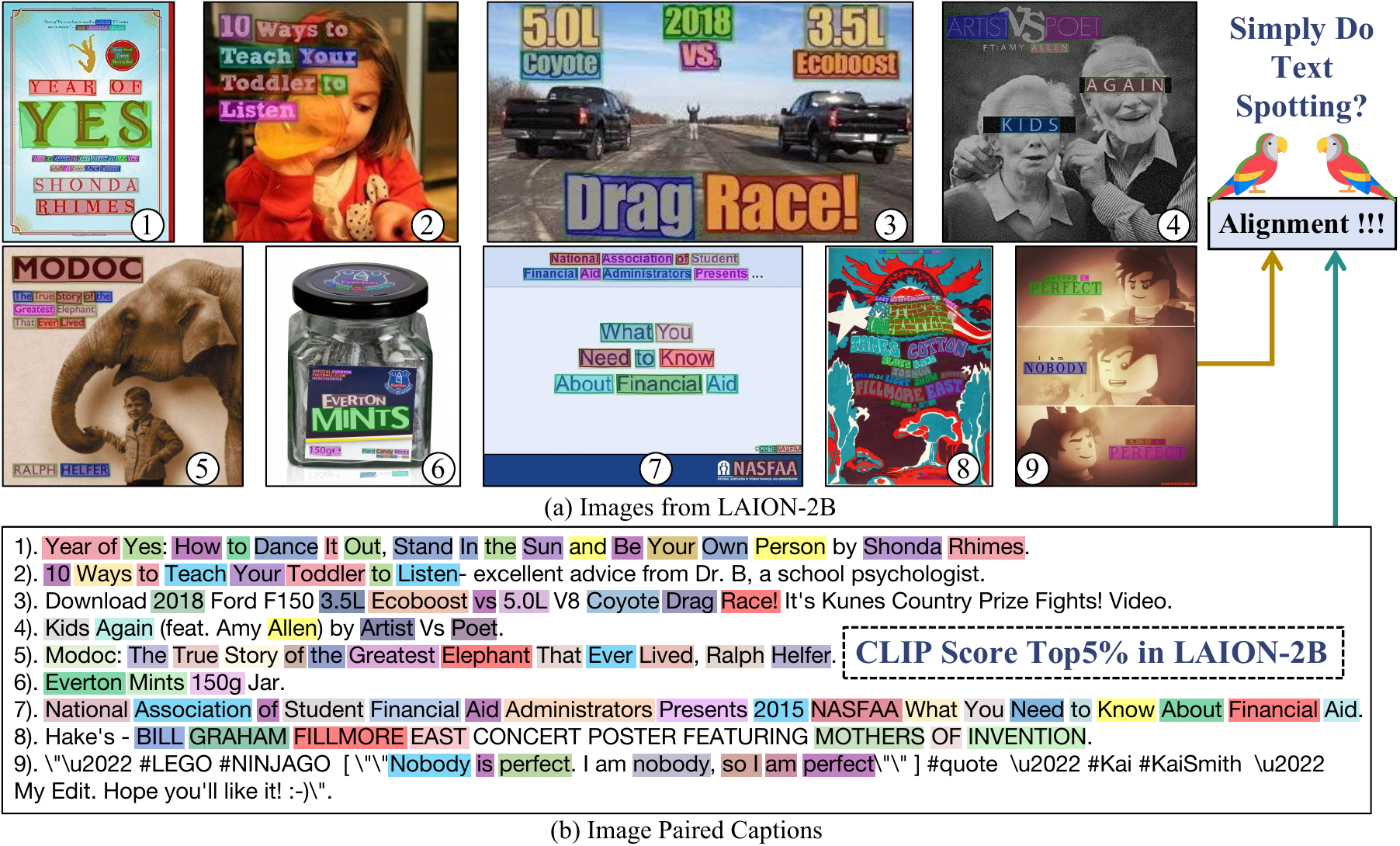}
    \vspace{-10pt}
    \captionof{figure}{\textbf{In LAION-2B~\cite{schuhmann2022laion}, image-text pairs with the Top-5\% highest similarity score are most dominant by visual text!}
    These samples have dense concurrent text appearing in captions and images (text form in pixels).
    We refer to their captions as \textbf{Parrot Captions} as they raise a question: \textit{Dose CLIP Simply Parroting Text in Images for Vision-Language Alignment?}
    The concurrent text is spotted by the OCR model and highlighted with color in image-text pairs. (Best view in color)
    }
\label{fig:teaser}
\end{center}
}]

\renewcommand{\thefootnote}{\fnsymbol{footnote}}
\footnotetext[1]{Equal contribution. \dag Corresponding author.}

\begin{abstract}
Despite CLIP being the foundation model in numerous vision-language applications, the CLIP suffers from a severe text spotting bias. Such bias causes CLIP models to `Parrot' the visual text embedded within images while disregarding the authentic visual semantics. We uncover that in the most popular image-text dataset LAION-2B, the captions also densely parrot (spell) the text embedded in images. Our analysis shows that around \textbf{50\%} of images are embedded with visual text content, and around \textbf{30\%} of captions words are in these embedded visual content. Based on such observation, we thoroughly inspect the different released versions of CLIP models and verify that the visual text is the dominant factor in measuring the LAION-style image-text similarity for these models. To examine whether these parrot captions shape the text spotting bias, we train a series of CLIP models with LAION subsets curated by different parrot-caption-oriented criteria. We show that training with parrot captions easily shapes such bias but harms the expected visual-language representation learning in CLIP models. This suggests that it is urgent to revisit either the design of CLIP-like models or the existing image-text dataset curation pipeline built on CLIP score filtering.
\end{abstract}    
\vspace{-10pt}
\section{Introduction}
\label{sec:intro}
Recently, contrastive learning models~\cite{radford2021learning,jia2021scaling,schuhmann2022laion} pre-trained with large-scale image-text pair data has led to numerous vision-language modeling task breakthroughs.
Due to its efficiency and simplicity, the pioneering work CLIP~\cite{radford2021learning} now serves as a foundation model in various applications~\cite{rombach2022high,nichol2021glide,li2022languagedriven,zhou2022learning}.
However, several works~\cite{goh2021multimodal,berg2022prompt} have shown that the CLIP models have perpetuating biases towards visual text~\cite{lemesle2022language,materzynska2022disentangling}, color~\cite{yao2021cpt,shtedritski2023does}, gender~\cite{wang2021gender}, etc.
In this paper, we focus on probing the visual text bias in CLIP, i.e., the capacity of spotting text in images.
Most of the previous cues~\cite{radford2021learning,materzynska2022disentangling,shtedritski2023does} attribute the sources of biases to the noisy pre-training data.
Therefore, we begin by taking a close look at the most popular dataset, LAION-2B~\cite{schuhmann2022laion}.

Considering the massive scale of the image-text data, it is non-trivial to assess the bias simply with a rough estimation.
To this end, we first do image clustering on the whole dataset and rank each cluster by the CLIP scores to analyze the most preferred types of image-text pairs under CLIP score measurement.
As shown in Fig.~\ref{fig:teaser}, we surprisingly observe that a decent number of samples with top CLIP scores have dense concurrent text appearing in the captions and the images in the form of pixels.
These samples break the assumption that the CLIP models leverage text supervision to align the visual and language concepts.
We refer to these captions as \textbf{Parrot Captions} as they provide another shortcut to achieve the same goal by teaching the CLIP to do text spotting even without perceiving the actual visual concepts.
To understand the underlying impact, we analyze the parrot captions from three perspectives: dataset, widely used released models, and model training.

Our main contributions are three-fold:

\begin{enumerate}
  \item  \textbf{Captions in LAION-2B have a significant bias towards describing visual text content embedded in the images.} 
  We provide thorough profiling using off-the-self text spotting models on the LAION-2B dataset and show that over 50\% of the images are embedded with visual text content.
  Moreover, by examining the spotted text content and the paired caption in each image-text pair, we find that over 90\% of the captions at least have one concurrent word and reach at least around 30\% words overlap between the caption and spotted text from images.
  \textit{This finding suggests that the basic assumption of image-text semantic alignment in CLIP does not stand its ground when training with LAION-style data.}
  
  \item \textbf{Released CLIP models have strong text spotting bias almost in every style of web images, resulting in the CLIP-filtering datasets inherently biased towards visual text dominant data.}
  We investigate the OpenAI released CLIP model's behaviors in the LAION-2B dataset by examining the difference between alignment scores before and after text removal.
  The results show that the CLIP model predictions densely correlate the visual text embedded in images with their parrot captions.
  Next, we further study the preference of the text spotting capacity on text content in the CLIP and OpenCLIP models.
  Note that the CLIP is trained on WIT-400M while the OpenCLIP uses the LAION-2B dataset.
  Therefore, we use synthetic images embedded with specific rendered text to avoid overfitting in OpenCLIP.
  Our analysis shows that the OpenCLIP model is more biased towards text spotting than the CLIP model.
  \textit{We believe that the parrot caption plays a lurking role in training these released CLIP models and is the source of text spotting capacity instead of emergence~\cite{wei2022emergent} in language models}.
  
  \item \textbf{CLIP models easily learn text spotting capacity from parrot captions while failing to connect the vision-language semantics, just like a text spotting parrot.}
  We sample different LAION-2B subsets curated by text-orientated criteria, including the embedded text ratio, the concurrent word ratios, and the relative CLIP score from text removal to train CLIP models under the same setting.
  The results show that using parrot captions data, the CLIP model can learn strong text spotting capacity but lose most of the zero-shot generalization ability on image-text downstream tasks.
  \textit{Lastly, we argue that the existing data curation pipeline built on the CLIP score and the contrastive fashion urgently needs to be re-examined by considering such hidden parrot captions.}
  
\end{enumerate}
\section{Related Work}

\subsection{Contrastive Vision-Language Pre-training} 
Modeling vision and language by aligning the embedding similarity between paired image-text data~\cite{radford2021learning,jia2021scaling,schuhmann2022laion} has shown great potential for transferable to downstream vision-language tasks.
The pre-training techniques mainly contain the vision encoder~\cite{he2016deep,dosovitskiy2020image} for image embedding encoding, text encoder~\cite{devlin2018bert} for text embedding modeling, and cross-modal contrastive learning~\cite{radford2021learning,jia2021scaling,li2021align,zhai2022lit} for learning a joint embedding space of vision and language.
The pioneering work CLIP~\cite{radford2021learning} leverages 400 million noisy image-text pairs to learn transferable visual representation from text supervision and show impressive zero-shot performance for various vision-language tasks.
Following the CLIP, several vision-language models such as ALIGN~\cite{jia2021scaling}, BASIC~\cite{pham2023combined}, and Open-CLIP~\cite{schuhmann2022laion} are proposed, and the CLIP models have been replicated on various
datasets including WIT~\cite{radford2021learning}, LAION~\cite{schuhmann2022laion}, COYO~\cite{kakaobrain2022coyo}, and DataComp~\cite{gadre2023datacomp}.
We mainly profile the LAION-2B~\cite{schuhmann2022laion} dataset due to its large scale and wide usage~\cite{nichol2021glide,rombach2022high} and two versions of pre-trained models, CLIP and OpenCLIP. 
Note that the 2 billion image-text pairs in the LAION-2B dataset are filtered by OpenAI released CLIP models, making the OpenCLIP connect to CLIP closely.

\subsection{Studying of CLIP Behaviors}
Despite the strong zero-shot and transferable performance of CLIP, the perpetuating biases~\cite{goh2021multimodal,wang2021gender,agarwal2021evaluating,yuksekgonul2022and,lemesle2022language} in CLIP are still not well investigated due to its large-scale noisy training data.
Much research~\cite{yao2021cpt,materzynska2022disentangling,shtedritski2023does,xu2023devil} focuses on revealing or enhancing the downstream performance with discovered bias in CLIP.
For example, colorful masks~\cite{yao2021cpt} or red circles~\cite{shtedritski2023does} applied to images can improve the zero-shot performance on visual grounding or keypoint localization tasks.
In studying visual text content bias, ~\cite{goh2021multimodal} shows the multimodal neurons of CLIP not only respond to visual content and the visual text embedded in the image. 
Another work~\cite{lemesle2022language} shows that image recognition in CLIP can be strongly dominated by the visual text embedded in the image.
To disentangle such bias,~\cite{materzynska2022disentangling} attempts to separate the text spotting representation in pre-trained CLIP by training representation projection.
Meanwhile, LoGoPrompt~\cite{shi2023logoprompt} enhances the classification performance by utilizing the visual text content as auxiliary prompts as input.
Also, CLIPPO~\cite{tschannen2023clippo} shows that directly aligning the image and synthetic images embedded with the captions can perform similarly to CLIP without a text-specific encoder.

\begin{figure}
\centering
\includegraphics[width=\linewidth]{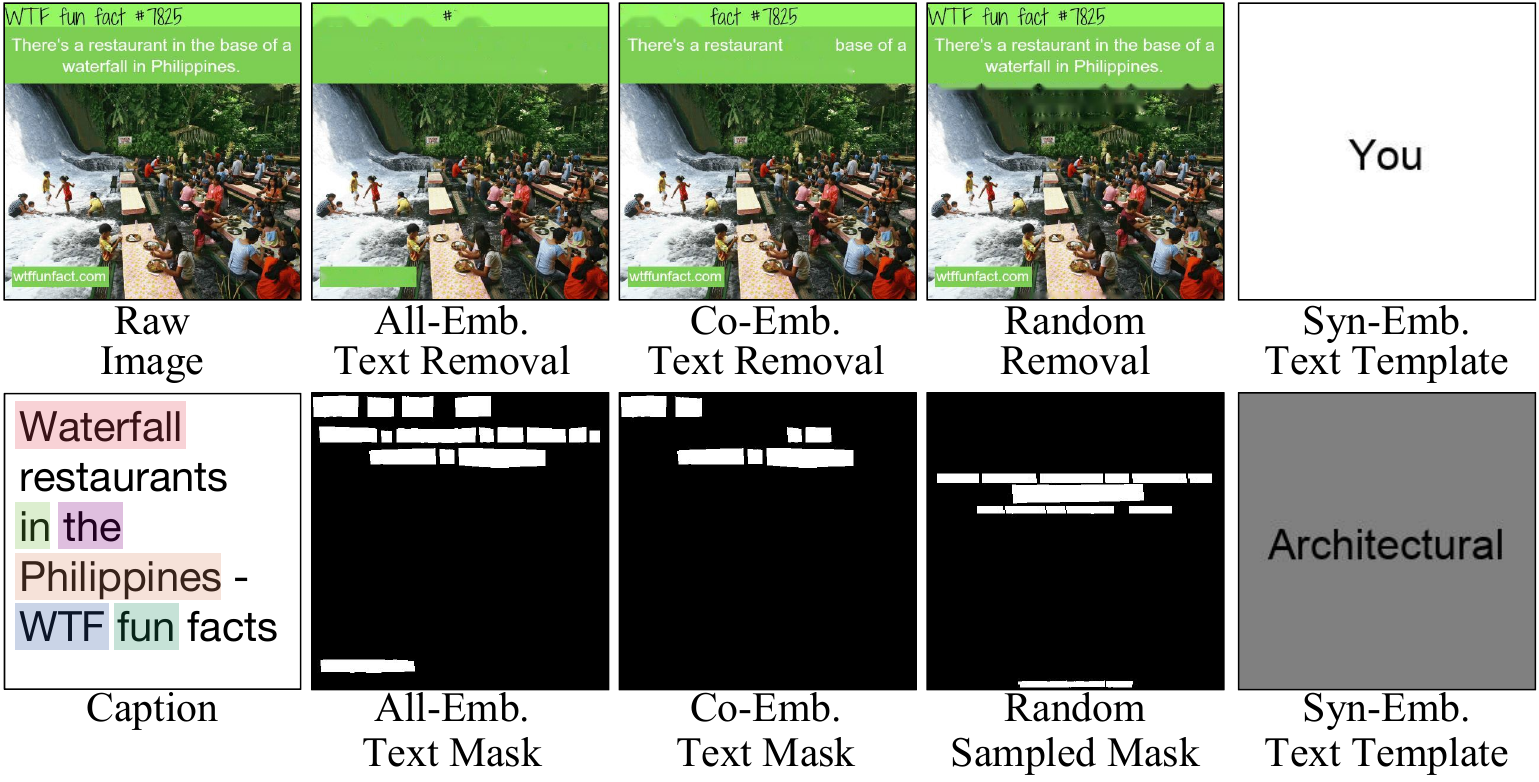}
\vspace{-18pt}
\caption{Visualization of defined terminologies. Co-emb.~text is highlighted in the caption with colors.
}
\vspace{-10pt}
\label{fig:term}
\end{figure}

\subsection{Data Curation with Text Removal}
Due to the successful practice of data curation in LAION datasets~\cite{laion400,schuhmann2022laion} on scaling up the image-text datasets, searching advanced selection strategy to improve the data quality from common crawl data pool gains a growing interest~\cite{gadre2023datacomp}.
Recently, several works~\cite{radenovic2023filtering,maini2023t,cao2023less} suggest that introducing text-related filtering methods improves the pre-training dataset quality.
In DiHT~\cite{radenovic2023filtering}, the data curation steps include filtering out the image-text pairs with high OCR confidence and matching text ratio.
Moreover,~\cite {maini2023t,cao2023less} mainly focus on studying the importance of filtering out the text-dominate images utilizing OCR models to improve pre-training dataset quality.
Maini et al.~\cite{maini2023t} also draw the observation that 40\% of LAION’s image text is highly correlated with the caption, but only performing a small pilot study on 500 samples with manual judgment.
\textit{Differently, this paper makes the first attempt to reveal the source of text spotting capacity in CLIP is the data bias and the consequences of such bias in existing commonly used datasets.}

\begin{algorithm}[t]
\caption{Pseudocode of Detecting Co-Emb.~Text (Rate)}
\label{alg:code}
\definecolor{codeblue}{rgb}{0.25,0.5,0.5}
\lstset{
  backgroundcolor=\color{white},
  basicstyle=\fontsize{7.2pt}{7.2pt}\ttfamily\selectfont,
  columns=fullflexible,
  breaklines=true,
  captionpos=b,
  commentstyle=\fontsize{7.2pt}{7.2pt}\color{codeblue},
  keywordstyle=\fontsize{7.2pt}{7.2pt},
}

\begin{lstlisting}[language=python][b]
# caption: captions from LAION-2B dataset.
# ocr_text: text spotted by OCR model.
cap_words = set(caption.split())
ocr_words = set(ocr_text.split())
co_emb_text = intersection(cap_words, ocr_words) 
co_emb_text_rate = len(co_emb_text) / len(cap_words)
\end{lstlisting}

\end{algorithm}

\section{Terminology}
The data processing on images in Sec.~\ref{sec:data},~\ref{sec:pretrained},~\ref{sec:training} mainly cover clustering, text spotting (OCR), and text inpainting.
Firstly, we cluster all images based on feature similarity.
For each image-text pair, we then use the pre-trained text spotting model to detect and recognize the text print in image pixels. 
The mask images in Fig.~\ref{fig:term} are the spotted text area.
Next, we match the spotted text with the caption using Algorithm~\ref{alg:code} to obtain the concurrent words and their ratio in captions.
Lastly, we use inpainting to remove the text from the image for the CLIPs' pattern ablation.
To avoid confusion, we define these concepts as follows,

\definecolor{fig_blue}{rgb}{0.53, 0.81, 0.92}
\definecolor{fig_red}{rgb}{0.95, 0.51, 0.50}
\definecolor{fig_green}{rgb}{0.56, 0.93, 0.56}

\begin{figure*}[tp]
    \centering
    \begin{subfigure}{\textwidth}
        \includegraphics[width=\linewidth]{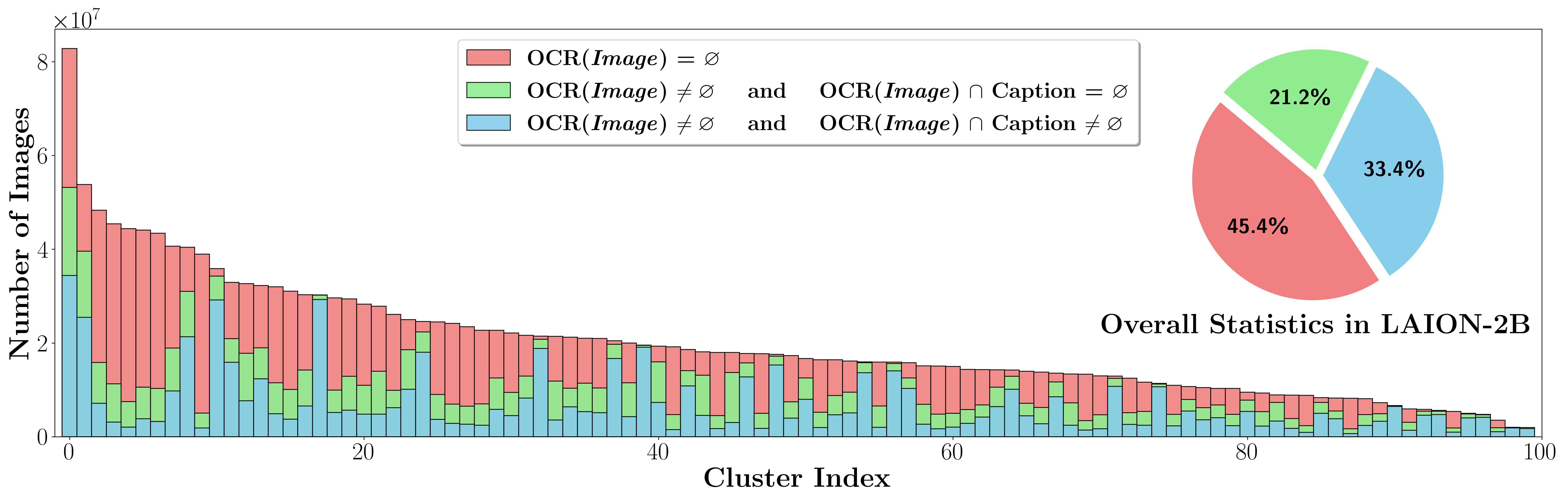}
        \vspace{-16pt}
        \caption{The ratio of different OCR-oriented data types in LAION-2B clusters.}
        
    \end{subfigure}
    \hfill
    \begin{subfigure}{\textwidth}  
        \includegraphics[width=\linewidth]{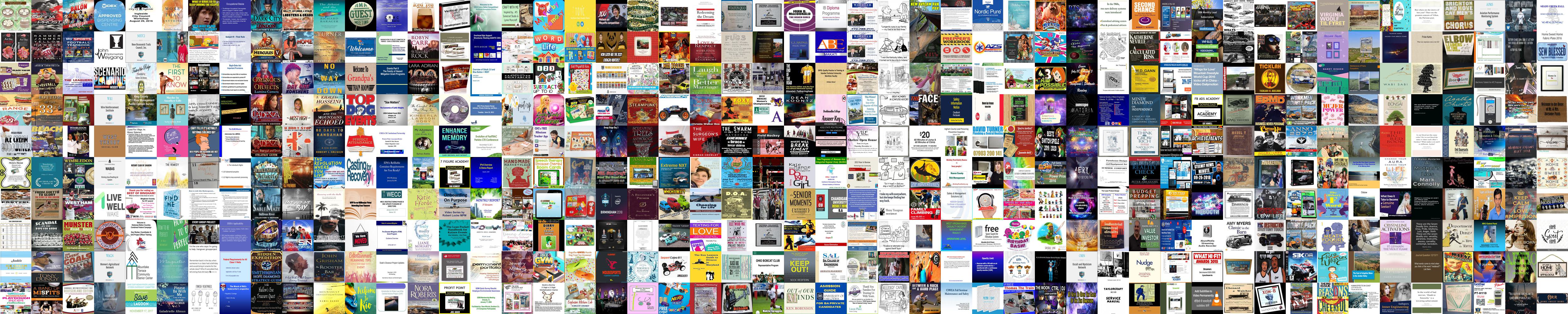}
        \vspace{-12pt}
        \caption{Top CLIP score samples visualization from 50 clusters with \raisebox{0.5ex}{\fcolorbox{black}{fig_blue}{\makebox(1,1){}}} ratio over 80\%.}
    \end{subfigure}
    \vspace{-15pt}
    \caption{\textbf{(a)}: Based on the OCR prediction results, the image-text pairs are divided into three types: \raisebox{0.5ex}{\fcolorbox{black}{fig_red}{\makebox(1,1){}}} image without visual embedded text content; \raisebox{0.5ex}{\fcolorbox{black}{fig_green}{\makebox(1,1){}}} the spotted text from the image has no concurrent text with the caption; \raisebox{0.5ex}{\fcolorbox{black}{fig_blue}{\makebox(1,1){}}} the spotted text at least share one concurrent word with the caption. 
    The clusters are merged from 4000 into 100 for a better view. \textbf{(b)}: In the clusters with high \raisebox{0.5ex}{\fcolorbox{black}{fig_blue}{\makebox(1,1){}}} ratio, the top CLIP score samples contain various text sources, such as posters, book covers, advertisements, TV show screenshots, and even PowerPoint slides.}
    \label{fig:overall_stat}
\end{figure*}

\begin{itemize}
    \item \textbf{Embedded Text}: text spotted by OCR models from the images.
    To study the correlation of embedded text with captions, we define different kinds of embedded text as,
    \begin{itemize}
        \item \textbf{All-Emb.~Text}: all the text is spotted from an image.
        \item \textbf{Co-Emb.~Text}: spotted text concurrently appears in the image's corresponding captions.
        \item \textbf{Syn-Emb.~Text}: synthetic text rendered in an image with a fixed font and a blank background.
    \end{itemize}
    Fig.~\ref{fig:term} shows examples of spotted embedded text by binary mask and the rendering results of synthetic text.
    \item \textbf{Co-Emb.~Text Rate (CoTR)}: the word set IoU of Co-Emb.~text and captions (See Algorithm.~\ref{alg:code}).
    \item \textbf{Parrot Caption}: captions with CoTR $> 0$.
    \item \textbf{Image w/ or w/o Embedded Text}: spotted text results of a given image are none-empty or empty.
    \item \textbf{Text Removal Image}: do inpainting in the specific spotted text area (All-Emb., Co-Emb., or Random).
    The random is implemented by sampling other image's text areas. 
    For the different inpainting results, see Fig.~\ref{fig:term}.
    \item \textbf{Relative Scores (RSA/RSC)}: the difference of the CLIP score between images modified by different inpainting operations while keeping the same captions.
    \textbf{RSA} and \textbf{RSC} are the short for the relative scores before and after removing All-Emb.~text and Co-Emb.~text.
    \item \textbf{Image Clusters}: image partitions based on K-Means.
    \item \textbf{CLIP and OpenCLIP}: the CLIP models are trained on WIT-400M~\cite{radford2021learning} and LAION-2B~\cite{schuhmann2022laion} dataset.
    \item \textbf{N-gram Vocabulary (Vocab)}: the set of all contiguous N word sequences extracted from a text corpus, such as the collection of all captions or embedded text.

\end{itemize}

\section{Profiling LAION-2B Data}
\label{sec:data}
To better profile the image-text pair data on a billion scale, we first cluster all the images based on CLIP features into 4,000 clusters and sort each cluster with CLIP scores.
After obtaining all the cluster labels, we use the SOTA text spotting model~\cite{ye2023deepsolo} to get the visual text content on all the collected images.
Finally, we aggregate all the model-predicted results and compare them with their corresponding captions to bring out our observations.

\subsection{Implementation Details}
\noindent\textbf{Clustering with CLIP Features:}
We first train K-Means (implemented by Faiss~\cite{johnson2019billion}) on the LAION-400M~\cite{laion400} subset using ViT-B-32~\cite{dosovitskiy2020image} CLIP features to speed up the clustering process.
Due to the large memory consumption, we reduce the feature dimensions from 512 to 256 using PCA (implemented by scikit-learn~\cite{scikit-learn}).
Then, we scan and partition the whole dataset using trained K-Means with the same feature extraction pipeline.

\noindent\textbf{Text Spotting and Matching:}
To detect and recognize text across various scenes, we adopt DeepSolo~\cite{ye2023deepsolo} as our text-spotting model and use the pre-trained checkpoints with the ViTAEv2-S~\cite{zhang2023vitaev2} backbone in default setting.
The output format of the text spotting model is a sequence of polygons of text location and their recognized characters.
Despite its strong performance, we empirically find that DeepSolo can not handle the crowd scenes well (with more than 100 separate words) but is only a small proportion of the dataset ($\sim$2\%).
To identify the correlation between the spotted text and captions, we use Algorithm~\ref{alg:code} to calculate the Co-Emb.~text rate in each image-text pair.
Considering the predictions that the text spotting model might miss or misspell words in complex scenes, we also use Levenshtein distance to calculate the fuzzing similarity and reported in Tab.~\ref{table:overall_stat}.

\begin{table}[t]
\centering
\begin{tabular}{l|c}
\toprule
Num. of Total Img. & 1,985,284,122 \\
Num. of Img. w/ Emb.~Text & 1,083,896,427 \\
Num. of Img. w/ Co-Emb.~Text & 663,600,432 \\
\hline
Co-Emb.~Text Rate (in Total) & 15.42\% \\
-- (in Img. w/ Emb.~Text) & 28.24\% \\
Fuzzy Co-Emb.~Text Rate (in Total) & 30.46\% \\
-- (in Img. w/ Emb.~Text) & 55.79\%  \\
\bottomrule
\end{tabular}
\vspace{-5pt}
\caption{
\textbf{Overall correlation statistic} between spotted text and captions in the LAION-2B.
More than 50\% of images are embedded with text, and 30\% of caption words are printed in images!
}
\label{table:overall_stat}
\vspace{-10pt}
\end{table}

\subsection{Statistic and Observations from LAION-2B}
The overall statistics of the 2 billion image-text pairs are reported in Tab.~\ref{table:overall_stat}.
In summary, the images embedded with visual text content reach a surprisingly high proportion of 54.60\% in the investigated data.
Around 15\% of words in the dataset captions are Co-Emb.~text, and the proportion of Co-Emb.~text can further reach 30\% when considering the fuzzy matching results of the spotted text and captions.
This suggests that the CLIP models trained on these data might lead to a high bias toward text spotting.

To better visualize the data distribution, we provide cluster-specific statics results and top CLIP score samples of text-dominated clusters in Fig.~\ref{fig:overall_stat}.
We divide all images into 100 clusters based on visual similarity and classify them into three data types according to the OCR results.
Every cluster contains more or less images embedded with text.
Combined with sample visualization, we observe that in the LAION collected data, the parrot captions cover various scenes.
In the subsets of images embedded with text, around 60\% of captions at least precisely parrot one concurrent word (Co-Emb. Text Rate $> 0$) appearing in the image.
It suggests that the data collection pipeline of LAION~\cite{schuhmann2022laion} has a strong bias to introduce parrot captions from web data.

\begin{figure}
    \centering
    \includegraphics[width=\linewidth]{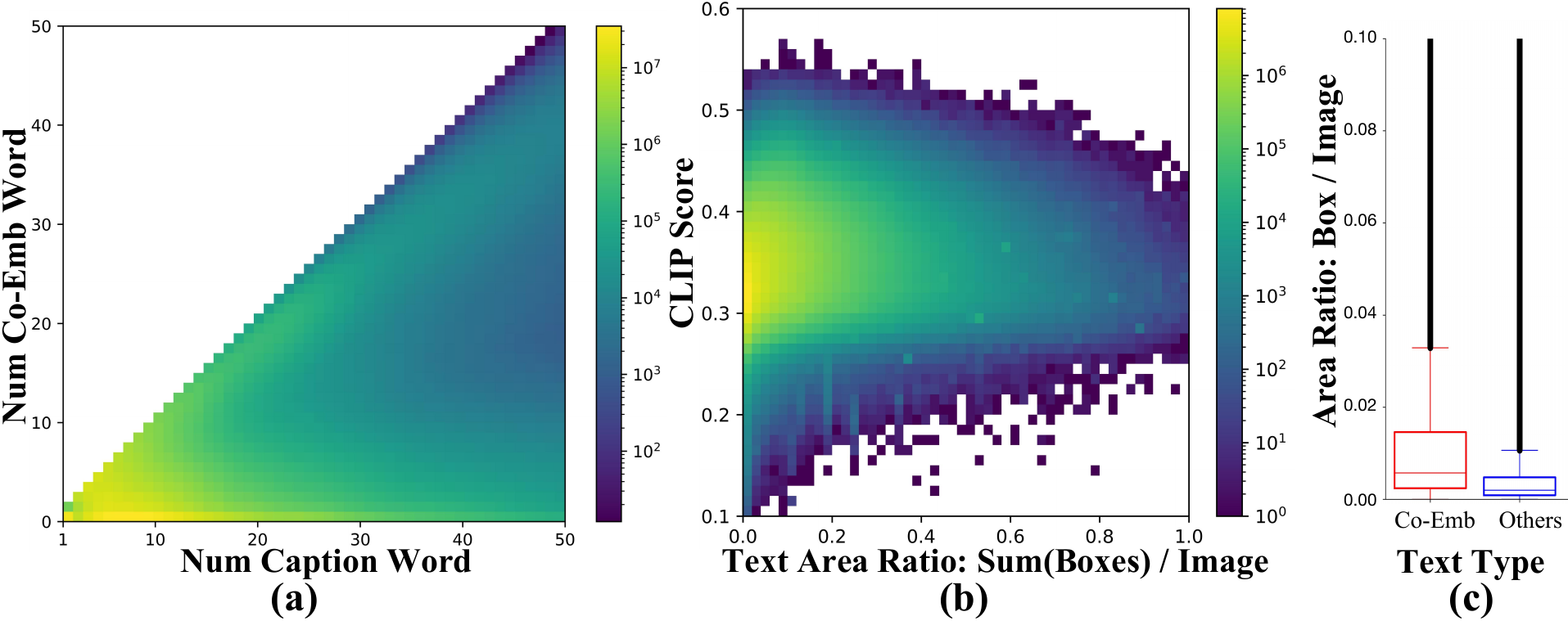}
    \vspace{-18pt}
    \caption{ 
    \textbf{(a)}: Visualization of the number of caption words and associated spotted concurrent words based on precise word matching. \textbf{(b):} Distribution of total area of concurrent words placed in the image and its ViT-B CLIP score. \textbf{(c):} Distribution of text size of the single concurrent word and other spotted word.}
    \vspace{-8pt}
    \label{fig:corr_count}
\end{figure}

To better understand Co-Emb.~Text, we provide a more thorough analysis of the word counting and text size of parrot captions.
As shown in Fig.~\ref{fig:corr_count}\textcolor{red}{a}, the results show that a large proportion of the Co-Emb.~Text only takes a few words.
However, we also find a large number of captions that are almost full parrot captions (see areas around the heatmap diagonal).
Next, in Fig.~\ref{fig:corr_count}\textcolor{red}{b}, we investigate the correlation between the size of concurrent words and CLIP score.
The results show that the large text size does not usually lead to a higher score; meanwhile, the small text size can also dominate the score.
One possible reason is the text content and input resolution may matter more for CLIP.
Moreover, we discover that the larger text is more likely to be parroted in the captions, as shown in Fig.~\ref{fig:corr_count}\textcolor{red}{c}.
\section{Inspecting Pre-Trained CLIP Models}
\label{sec:pretrained}
It is important to note that the LAION-2B dataset collection pipeline uses the CLIP score from OpenAI's model to filter out the image-text pair below \textbf{0.28}.
Therefore, we inspect these two released CLIP models~\cite{radford2021learning,schuhmann2022laion} to answer better why LAION data contains such a high proportion of parrot captions.
Specifically, the OpenAI's CLIP model is trained on the WIT dataset (out-of-domain model), and the OpenCLIP is trained on LAION-2B (in-domain model).
We first study whether the embedded text is the key factor in CLIP filtering by ablating the embedded text using text inpainting.
Moreover, we further investigate whether the text spotting capacity prefers specific text content by examining synthetic images with Syn-Emb.~text.

\begin{figure*}[t]
\centering
\includegraphics[width=\textwidth]{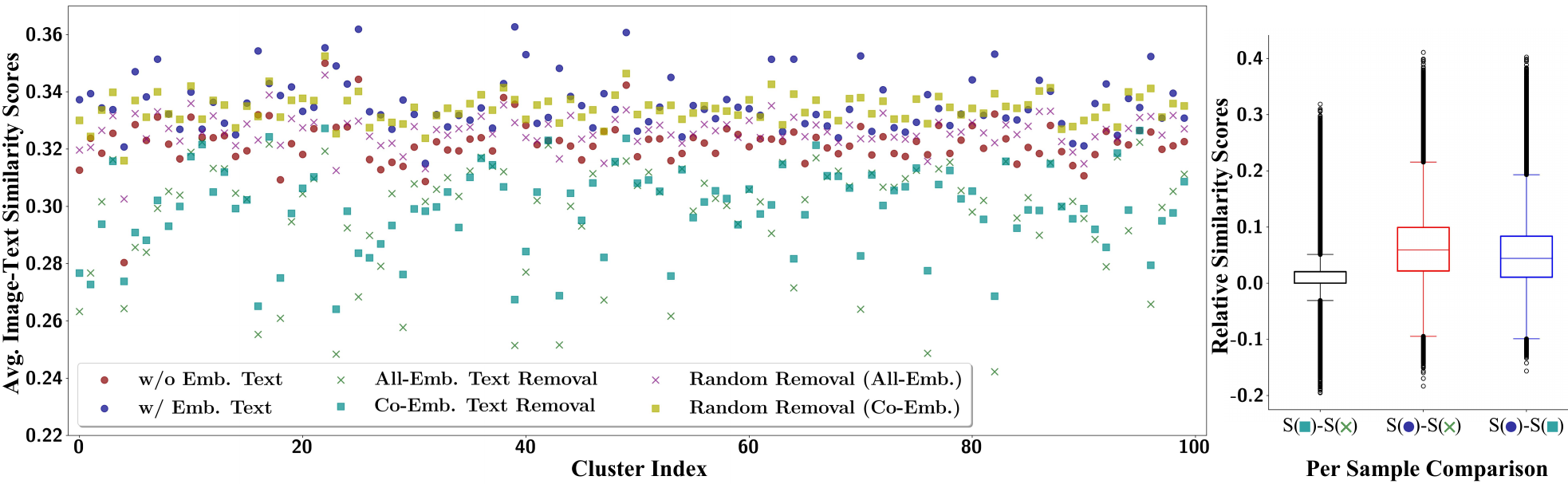}
\vspace{-20pt}
\caption{\textbf{Left:} Mean CLIP scores of image-text pairs with different text removal operations depicted in Sec~\ref{subsec:ab_remove}, and the data are grouped by cluster the same as Fig.~\ref{fig:overall_stat}. \textbf{Right:} Overall relative CLIP score distribution by comparing different text removal operations.
}
\label{fig:score_distribution}
\end{figure*}


\subsection{Ablation of Embedded Text Removal}
\label{subsec:ab_remove}
\noindent\textbf{Text Removal via Inpainting:}
Given the OCR predicted results, we use the fast marching method~\cite{telea2004image} to do text in-painting in the area of the spotted text polygons.
Accordingly, we generate two versions of text removal results for each image with embedded text, i.e., All-Emb.~text removal and Co-Emb.~text removal, as the parrot caption prediction is imperfect due to the limitation of OCR models.
We also generate random inpainting images with randomly sampled spotted text polygons from other images to ablate the information loss caused by image inpainting.
The spotted text masks and inpainting results are shown in Fig.~\ref{fig:term}.

\noindent\textbf{Results:}
Based on the OCR predicted results and text inpainting operations, we can obtain six types of LAION images, including 
{\color{wo_text_mark}{\large$\bullet$}}): images without any embedded text (OCR results are empty); 
{\color{raw_mark}{\large$\bullet$}}): images with any embedded text (OCR results are none-empty);
{\color{all_mark}{{$\boldsymbol{\times}$}}}): images removed All-Emb.~text (Inpaint all the areas of OCR predicted text);
{\color{co_mark}{\scriptsize$\blacksquare$}}): images removed Co-Emb.~text (Inpaint the areas of concurrent text in OCR predicted text and captions);
{\color{rand_all_mark}\textbf{{$\boldsymbol{\times}$}}}): images with random inpainting by other image's All-Emb.~text area, and
{\color{rand_co_mark}{\scriptsize$\blacksquare$}}): images randomly inpainted by other image's Co-Emb.~text area.
Then, we calculate the CLIP scores of all the groups of images and their paired captions using OpenAI released CLIP model (ViT-B-32). Fig.~\ref{fig:score_distribution} reports the mean scores of different types of images in each cluster and raises four observations as follows,

\begin{table}[]
\centering
\scalebox{0.9}{
\begin{tabular}{@{}l|c@{}}
\toprule
Setup & CLIP Score \\ \hline
{\color{wo_text_mark}{\large$\bullet$}} Raw w/o Emb. Text & 0.3223 $\pm$ 0.0078 \\
{\color{raw_mark}{\large$\bullet$}} Raw w/ Emb. Text & 0.3358 $\pm$ 0.0094 \\ \hline
{\color{rand_all_mark}\textbf{\scriptsize{$\boldsymbol{\times}$}}} Random All-Emb. Text Removal& 0.3260 $\pm$ 0.0057 \\
{\color{all_mark}{\scriptsize{$\boldsymbol{\times}$}}} All-Emb. Text Removal & 0.2974 $\pm$ 0.0197 \\ \hline
{\color{rand_co_mark}{\scriptsize$\blacksquare$}} Random Co-Emb. Text Removal & 0.3341 $\pm$ 0.0051\\
{\color{co_mark}{\scriptsize$\blacksquare$}} Co-Emb. Text Removal & 0.2993 $\pm$ 0.0146\\ \bottomrule
\end{tabular}}
\vspace{-8pt}
\caption{
\textbf{Mean CLIP score of different setups of text removal.}
}
\label{tab:distribution}
\vspace{-8pt}
\end{table}

\begin{itemize}
    \item The images embedded with text achieve higher CLIP scores in most clusters than those without embedded text.
    \item The CLIP scores significantly drop once we remove the text from the images compared to its random inpainting baseline. It indicates that the parrot captions correlate highly with the CLIP score measurement. 
    \item The text spotting mechanism of CLIP might be similar to Bags-of-Words~\cite{yuksekgonul2022and}.
    Most of the relative CLIP scores (S({\color{co_mark}{\scriptsize$\blacksquare$}}) - S({\color{all_mark}{{$\boldsymbol{\times}$}}})) between images removed Co-Emb.~text and All-Emb.~text are positive, as shown in the right of Fig.~\ref{fig:score_distribution}.
    The straightforward reason is the images lose more visual information due to the larger in-painting area, while another possible reason is the imperfect text spotting prediction or the corner cases in the matching algorithm leaking parts of the concurrent text in images.
    \item Not all the samples are dominated by the embedded text, as some samples achieve higher scores after removing text, indicating the embedded text also can be a distractor.
\end{itemize}

\noindent\textbf{Discussion:}
Due to the text removal, the image distribution may shift from the CLIP training set.
Therefore, we provide two random removal baselines to examine the effect of distribution shift.
In Tab.~\ref{tab:distribution}, we report the mean scores of different setups.
Results show that the random baselines are very close to the raw image baseline, indicating that the CLIP model is robust to the distribution shift caused by information loss in inpainted regions.

\begin{figure}[]
    \centering
    \begin{subfigure}{0.85\linewidth}
        \includegraphics[width=\linewidth]{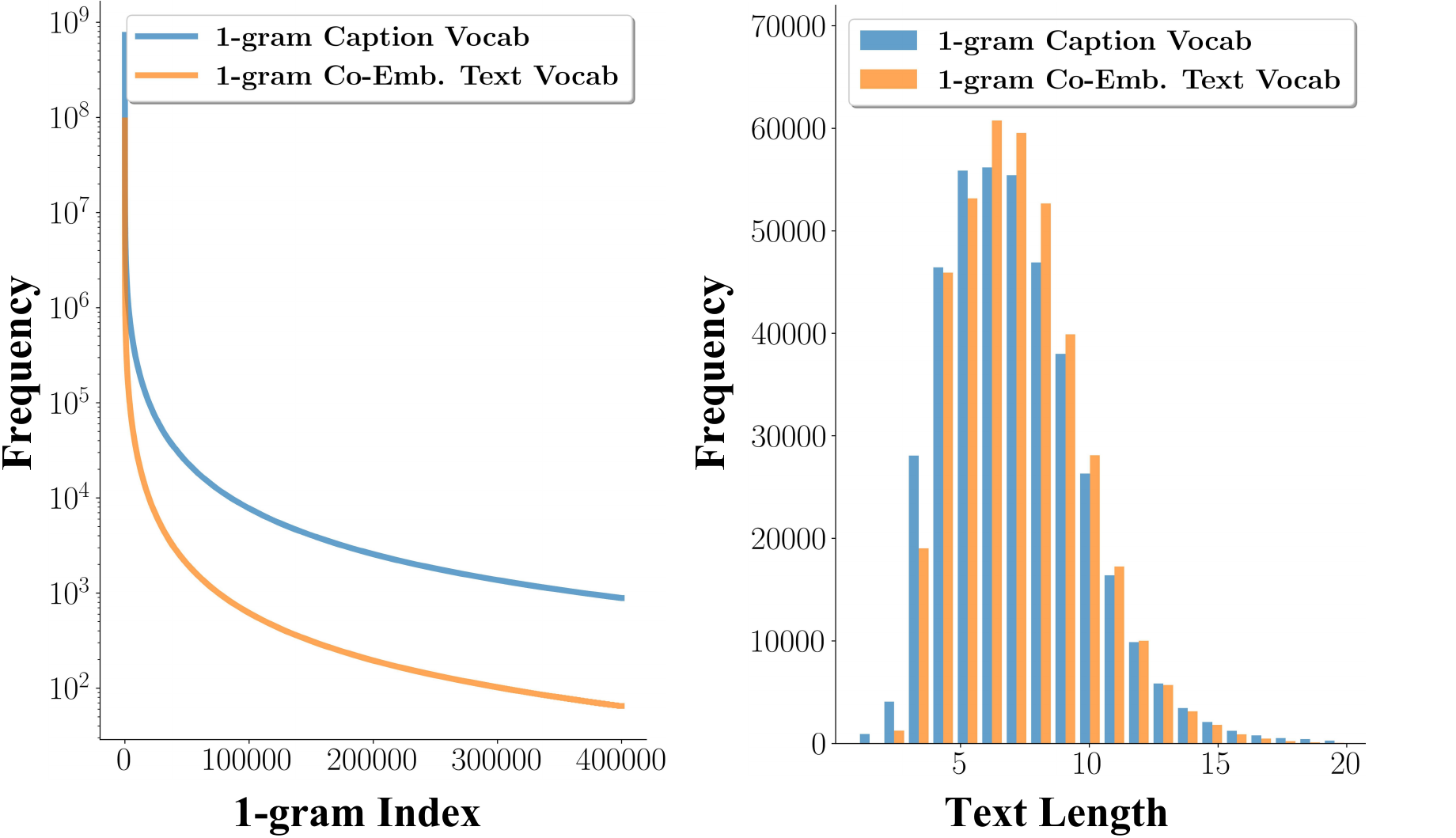}
        \caption{Statistic of 1-gram vocabularies.}
        \label{fig:vocab_stat}
    \end{subfigure}
    \begin{subfigure}{0.9\linewidth}
        \includegraphics[width=\linewidth]{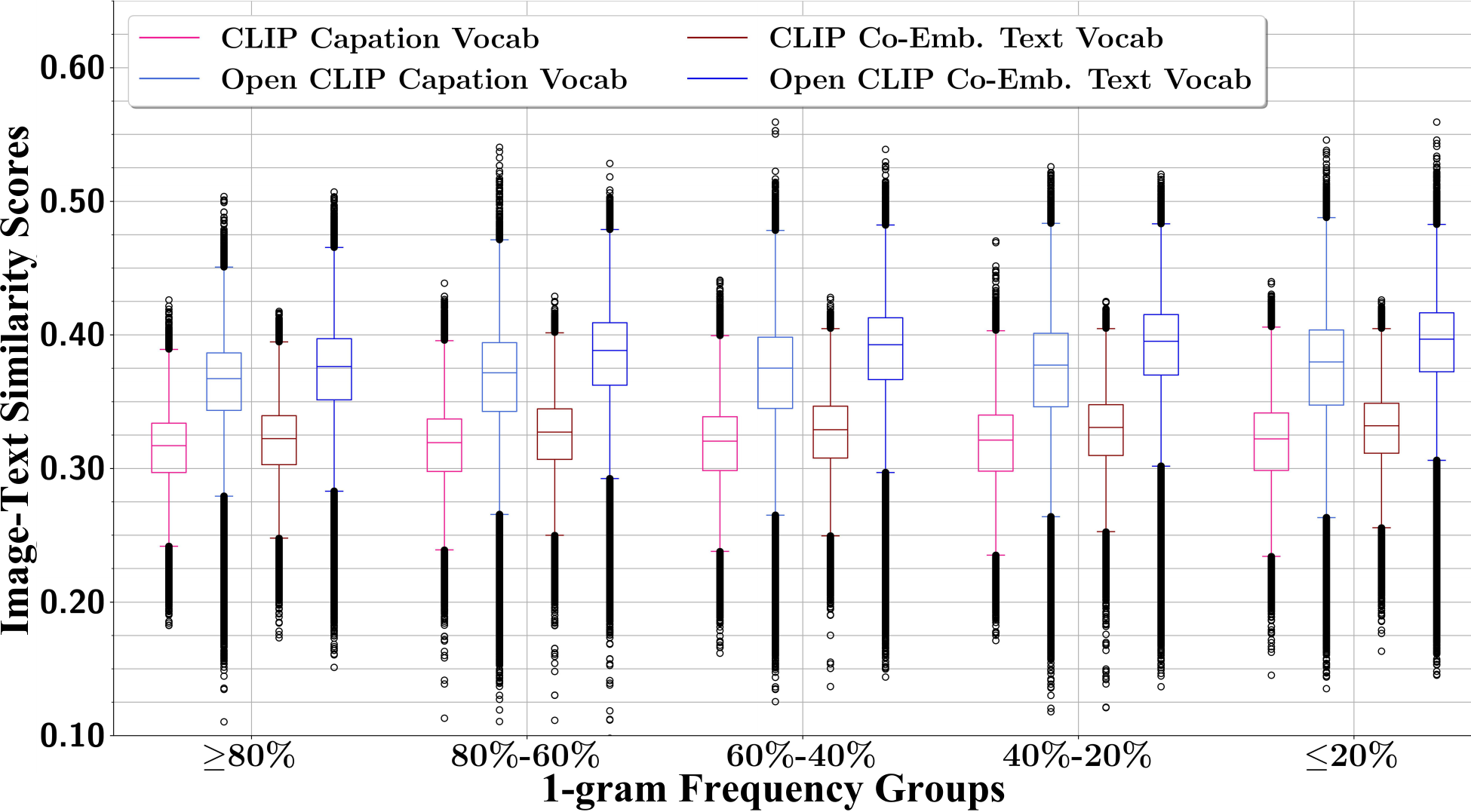}
        \caption{Grouped by 1-gram frequency intervals.}
        \label{fig:fre_group}
    \end{subfigure}
    \vspace{-8pt}
    \caption{
    \textbf{OpenCLIP more bias than the CLIP model.}
    Grouped score distributions of prompting CLIP and OpenCLIP models with N-gram Syn-Emb.~text and synthetic images.}
    \label{fig:figure}
\end{figure}

\subsection{Prompting with Syn-Emb.~Text}
\label{sec:ab_syn}
\noindent\textbf{Generating Synthetic Images from N-gram Vocabulary:} 
To investigate the CLIP models' text spotting preference, we adopt a similar strategy in~\cite{materzynska2022disentangling} to use synthetic images to embed specific text content by rendering text in a blank background.
For each text, we use four fore-background style rendering templates (black-white, black-grey, white-grey, and white-black), as shown in Fig.~\ref{fig:term}.
Different from the uniformly sampling letters in~\cite{materzynska2022disentangling}, we generate the text content from the N-gram vocabulary built from captions and Co-Emb.~text to study the text spotting pattern.
We only select the top frequent 400,000 grams for each vocabulary.
The statistics of 1-gram vocabulary are shown in Fig.~\ref{fig:vocab_stat}, which is a long-tail distribution.
Next, we calculate the synthetic images and their rendered text similarity on released ViT-B-32 CLIP and OpenCLIP models.

\begin{figure}[]
    \centering
    \begin{subfigure}{0.9\linewidth}  
        \includegraphics[width=\linewidth]{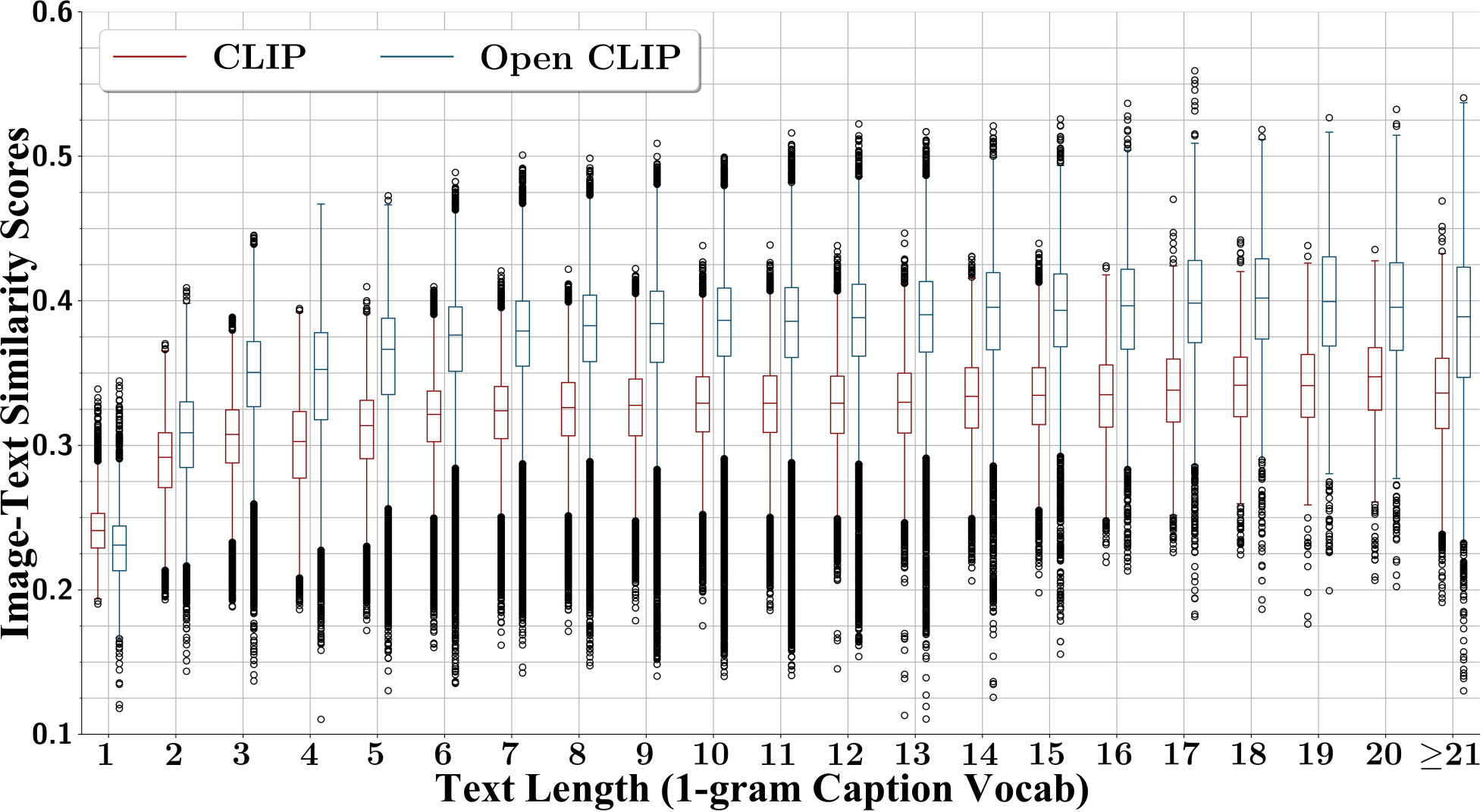}
        \caption{Grouped by Caption 1-gram length.}
        \label{fig:1_gram_group}
    \end{subfigure}
    \begin{subfigure}{0.9\linewidth}  
        \includegraphics[width=\linewidth]{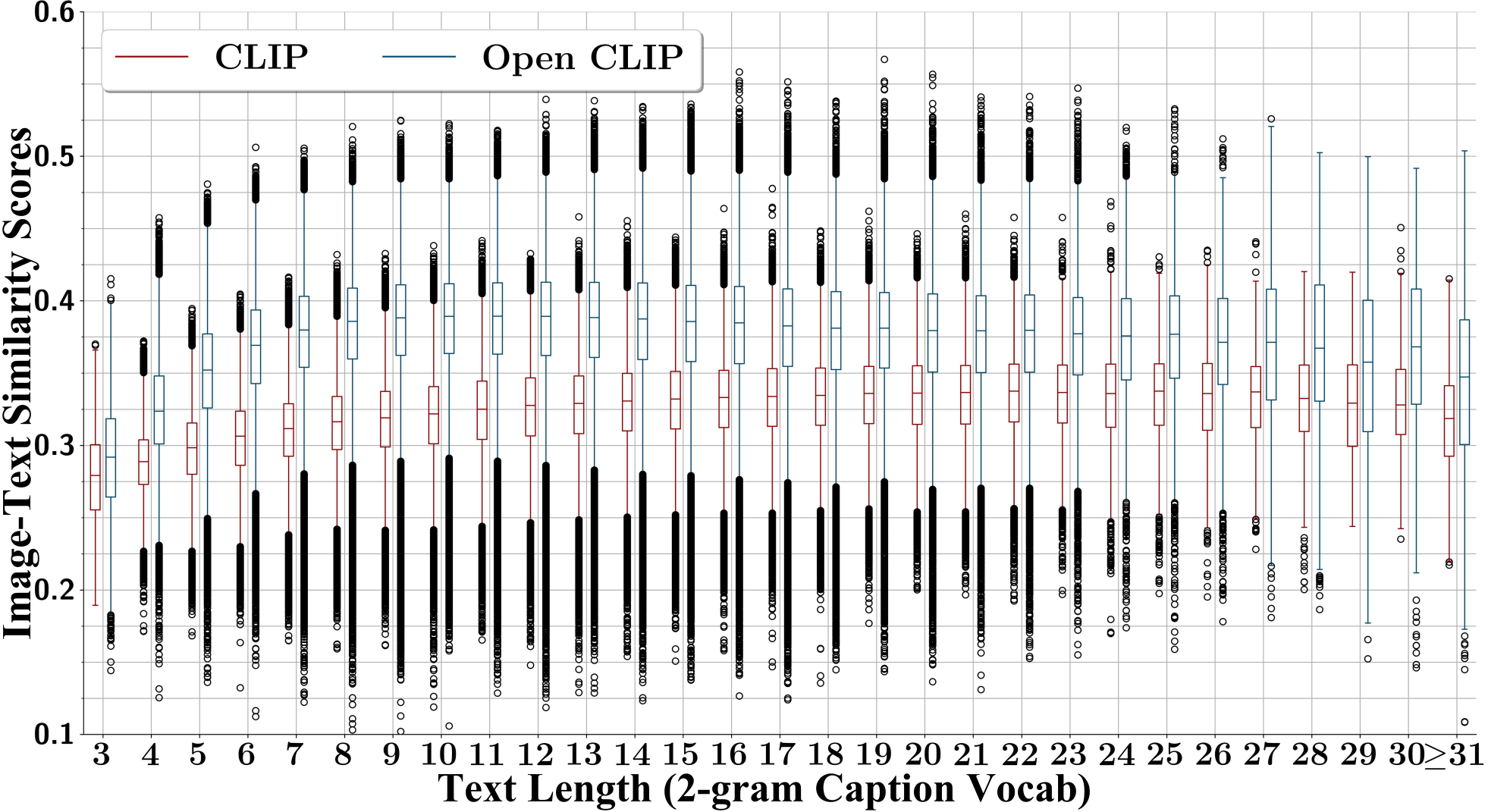}
        \caption{Grouped by Caption 2-gram length.}
        \label{fig:2_gram_group}
    \end{subfigure}
    \vspace{-8pt}
    \caption{
    \textbf{CLIPs are better at spotting the longer words.} Score distributions of N-gram Syn-Emb.~text grouped by text length.}
    \label{fig:figure}
\end{figure}

\noindent\textbf{Results:}
Firstly, we examine whether the CLIP models prefer recognizing more commonly seen words (with high frequency in vocabulary).
Therefore, we group the 1-gram results based on their frequency interval in the whole vocabulary, as shown in Fig.~\ref{fig:fre_group}.
The results show that the OpenCLIP model clearly has a stronger text spotting capacity than CLIP, i.e., more biased towards text spotting.
We also observe that all the CLIP models are more sensitive to the vocabulary built from the concurrent words.
Interestingly, both CLIP and OpenCLIP models have slightly higher scores on the less frequent grams.
Secondly, considering the long-tail grams might contain more characters, we further group the 1-gram and 2-gram results based on their text length in Fig.~\ref{fig:1_gram_group} and Fig.~\ref{fig:2_gram_group}.
Note that the Co-Emb.~text is not regularly arranged in the images, making it hard to extract continuous word sequences.
Results show that all the models are better at spotting the longer words, possibly due to the tokenizer used in the text encoder, making them more discriminative.
Meanwhile, in the groups of 2-gram samples, the scores gradually drop when spotting the extremely long text, indicating that the spotting capacity of CLIP models is possibly built on word-by-word.

\section{Training CLIP on Emb. Text Curated Data}
\label{sec:training}
Next, we dive deeper into the parrot captions by training CLIP models on LAION-2B subsets selected by different embedded-text-oriented criteria under the same setting.

\subsection{Experiment Setups}

\noindent \textbf{Training Setting and Implementation Details:}
We use the open-source software OpenCLIP~\cite{ilharco_gabriel_2021_5143773} for all the CLIP model training.
Our experiments are conducted on ViT-B~\cite{dosovitskiy2020image} and RN50~\cite{he2016deep} using 8 NVIDIA A100 GPUs for training.
We use 4,096 batch size for 3M and 8,192 for 12M scale subsets.
Other settings remain the same as~\cite{schuhmann2022laion}.

\begin{table}[tp]
    \centering
\scalebox{0.9}
{
    \begin{tabular}{l|c|ccc}
    \toprule
    Data & Model &  IN & Ret. & Avg. \\ \hline

    3M Random & RN50         & 0.204 & 0.222 & 0.256 \\
    3M w/o Emb.~Text & RN50   & 0.228\cellcolor{tabfirst} & 0.232\cellcolor{tabfirst} & 0.282\cellcolor{tabfirst} \\
    3M w/ Emb.~Text Only & RN50      & 0.071 & 0.139 & 0.164 \\
    \hline
    3M Random & ViT-B & 0.131 & 0.148 & 0.210 \\
    3M w/o Emb.~Text & ViT-B  & 0.162\cellcolor{tabfirst} & 0.164\cellcolor{tabfirst} & 0.234\cellcolor{tabfirst} \\
    3M w/ Emb.~Text Only & ViT-B     & 0.052 & 0.111 & 0.153 \\
    \hline

    12M Random & RN50        & 0.360 & 0.354 & 0.354 \\
    12M w/o Emb.~Text & RN50  & 0.409\cellcolor{tabfirst} & 0.361\cellcolor{tabfirst} & 0.372\cellcolor{tabfirst} \\
    12M w/ Emb.~Text Only & RN50     & 0.129 & 0.192 & 0.218 \\
    \hline
    
    12M Random & ViT-B      & 0.314 & 0.299 & 0.351 \\
    12M w/o Emb.~Text & ViT-B & 0.370\cellcolor{tabfirst} & 0.318\cellcolor{tabfirst} & 0.364\cellcolor{tabfirst} \\
    12M w/ Emb.~Text Only & ViT-B    & 0.129 & 0.172 & 0.225 \\
    \bottomrule
    \end{tabular}
    }
    \caption{\textbf{Ablation of images embedded with or without text.} The model trained on data without embedded text performs best on all tasks, while the data with embedded text damages the generalization capacity of learned representations.
    }
    \label{tab:ab_embtext}
\end{table}

\begin{table}[tp]
    \centering 
\scalebox{0.9}{
    \begin{tabular}{l|c|ccc}
    \toprule
    Data (3M) & Model &  IN & Ret. & Avg. \\ \hline
    CoTR = 0.0 & RN50       & \cellcolor{tabfirst}0.193 & \cellcolor{tabfirst}0.229 & \cellcolor{tabfirst}0.247 \\
    CoTR  $\geq$ 0.3 & RN50      & 0.031 & 0.110 & 0.137 \\
    CoTR  $\geq$ 0.5 & RN50      & 0.021 & 0.099 & 0.124 \\
    CoTR  $\geq$ 0.8 & RN50      & 0.012 & 0.082 & 0.096 \\
    CoTR  = 1.0 & RN50     & 0.012 & 0.074 & 0.102 \\
    \hline
    CoTR  = 0.0 & ViT-B   & \cellcolor{tabfirst}0.132 & \cellcolor{tabfirst}0.164 & \cellcolor{tabfirst}0.206 \\
    CoTR  $\geq$ 0.3 & ViT-B  & 0.029 & 0.084 & 0.130 \\
    CoTR  $\geq$ 0.5 & ViT-B  & 0.021 & 0.082 & 0.119 \\
    CoTR  $\geq$ 0.8 & ViT-B  & 0.012 & 0.076 & 0.104 \\
    CoTR  = 1.0 & ViT-B & 0.013 & 0.076 & 0.103 \\
    \bottomrule
    \end{tabular}}
    \caption{\textbf{Ablation of different Co-Emb.~Text Rate(CoTR).}
    The fewer parrot captions, the better downstream task performance.
    }
    \label{tab:3m_cotr}
\end{table}

\begin{table}[tp]
\scalebox{0.9}{
    \centering
    \begin{tabular}{l|c|c|ccc}
    \toprule
    Data (3M) & Model  & Avg.S({\color{raw_mark}{\large$\bullet$}})&  IN & Ret. & Avg. \\ \hline
    RSA $<$ 0.0& RN50  &0.319 & \cellcolor{tabfirst}0.181 & \cellcolor{tabfirst}0.220 & \cellcolor{tabfirst}0.239 \\
    RSA $\geq$ 0.0& RN50 &0.339 & 0.126 & 0.180 & 0.215 \\
    RSA $\geq$ 0.1& RN50 &0.351 & 0.041 & 0.123 & 0.148 \\
    RSA $\geq$ 0.2& RN50 &0.360 & 0.017 & 0.094 & 0.109 \\
    RSA $\geq$ 0.3& RN50 &0.376 & 0.009 & 0.075 & 0.097 \\
    \hline
    RSA $<$ 0.0& ViT-B &0.319 & \cellcolor{tabfirst}0.123 & \cellcolor{tabfirst}0.159 & \cellcolor{tabfirst}0.198 \\
    RSA $\geq$ 0.0& ViT-B &0.339 & 0.079 & 0.129 & 0.185 \\
    RSA $\geq$ 0.1& ViT-B &0.351 & 0.031 & 0.103 & 0.134 \\
    RSA $\geq$ 0.2& ViT-B &0.360 & 0.012 & 0.080 & 0.103 \\
    RSA $\geq$ 0.3& ViT-B &0.376 & 0.006 & 0.070 & 0.096 \\
    \bottomrule
    \end{tabular}
    }
    \caption{\textbf{Ablation of models trained on subsets sampled by different RSA}. RSA denotes the relative similarity (S({\color{raw_mark}{\large$\bullet$}}) $\mathbf{-}$ S({\color{all_mark}\textbf{{$\times$}}})) of raw S({\color{raw_mark}{\large$\bullet$}}) and removed All-Emb.~text S({\color{all_mark}\textbf{{$\times$}}}) images.
    }
    \label{tab:ab_rsa}
\end{table}

\begin{table}[tp]
\scalebox{0.9}{
    \centering
    \begin{tabular}{l|c|c|ccc}
    \toprule
    Data (3M) & Model & Avg.S({\color{raw_mark}{\large$\bullet$}}) & IN & Ret. & Avg. \\ \hline 
    RSC $<$ 0.0& RN50 &0.326 & \cellcolor{tabfirst}0.125 & \cellcolor{tabfirst}0.171 & \cellcolor{tabfirst}0.209 \\
    RSC $\geq$ 0.0& RN50 &0.345 & 0.062 & 0.129 & 0.168 \\
    RSC $\geq$ 0.1&RN50 & 0.354 & 0.014 & 0.091 & 0.106 \\
    RSC $\geq$ 0.2& RN50 &0.364 & 0.008 & 0.084 & 0.104 \\
    RSC $\geq$ 0.3& RN50 &0.380 & 0.005 & 0.058 & 0.084 \\
    \hline
    RSC $<$ 0.0& ViT-B &0.326 & \cellcolor{tabfirst}0.079 & \cellcolor{tabfirst}0.129 & \cellcolor{tabfirst}0.174 \\
    RSC $\geq$ 0.0& ViT-B &0.345 & 0.045 & 0.119 & 0.149 \\
    RSC $\geq$ 0.1& ViT-B &0.354 & 0.018 & 0.091 & 0.116 \\
    RSC $\geq$ 0.2& ViT-B &0.364 & 0.008 & 0.076 & 0.106 \\
    RSC $\geq$ 0.3& ViT-B &0.380 & 0.004 & 0.059 & 0.091\\
    \bottomrule
    \end{tabular}
    }
    \caption{\textbf{Ablation of models trained on subsets sampled by different RSC.}
    RSC denotes the relative similarity (S({\color{raw_mark}{\large$\bullet$}}) $\mathbf{-}$ S({\color{co_mark}{\scriptsize$\blacksquare$}})) of raw S({\color{raw_mark}{\large$\bullet$}}) and removed Co-Emb.~text S({\color{co_mark}{\scriptsize$\blacksquare$}}) images.
    }
    
    \label{tab:ab_rsc}
\end{table}

\begin{figure*}[tp]
\centering
\includegraphics[width=\textwidth]{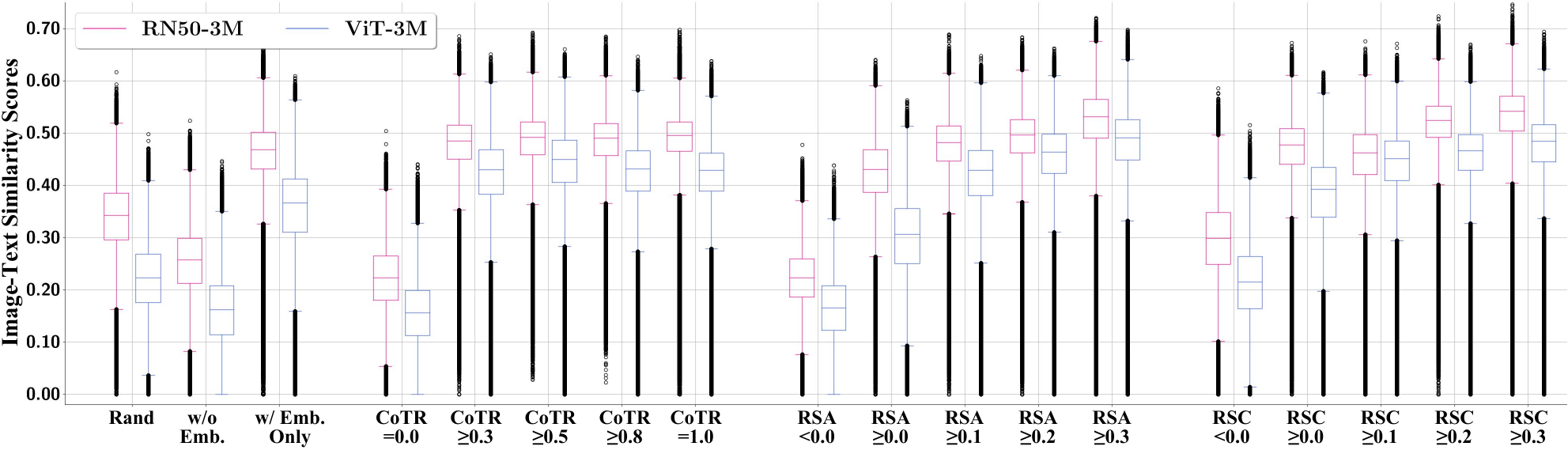}
\vspace{-18pt}
\caption{\textbf{CLIP models learn text spotting well from parrot captions.}
Benchmarking text spotting capacity of CLIP models with 1-gram caption vocabulary synthetic images dataset as the same as Sec.~\ref{sec:ab_syn}.}
\label{fig:subset_syn}
\end{figure*}

\noindent \textbf{Evaluation:}
We follow the DataComp benchmark~\cite{gadre2023datacomp} using 38 zero-shot classification and retrieval tasks as evaluation.
We report the average performance (Avg.) of the DataComp benchmark and two subset track performances, ImageNet (IN) and Retrieval (Ret.).
To evaluate the text spotting capacity, we use a synthetic benchmark the same as in Sec.~\ref{sec:ab_syn} and a real-world benchmark sampled from LAION-2B as the validation set.
In the synthetic benchmark, we calculate the similarity of all the 1-gram synthetic image-text pairs from caption vocabulary and report all the trained model results in Fig~\ref{fig:subset_syn}.
For the real-world benchmark, we sample 1M image-text pairs without any embedded text and 1M samples dominated by the parrot caption (the relative scores between raw and Co-Emb.~text removal images higher than 0.2).
Fig.~\ref{fig:subset_real} aggregates the mean scores of the 2M evaluation set and also reports the mean scores of applying text removal on the 2M evaluation set results.

\subsection{Ablation Study on Data Curation}
\label{sec:ab_std}
\noindent \textbf{Curation I: Embedded Text in Images.}
To study the impact of embedded text on overall pre-train data quality, we sample three subsets: random baseline, images without any embedded text, and images all embedded with text from LAION-2B.
The subsets include 3M and 12M scales.
The results in Tab.~\ref{tab:ab_embtext} show that images embedded with text generally reduce the pre-training dataset quality as all performance tracks significantly decrease.
Meanwhile, in Fig.~\ref{fig:subset_syn}, the model trained with the images embedded with text achieves the strongest text spotting capacity compared to the random and images without embedded text baselines.

\begin{figure}[tp]
\centering
\includegraphics[width=\linewidth]{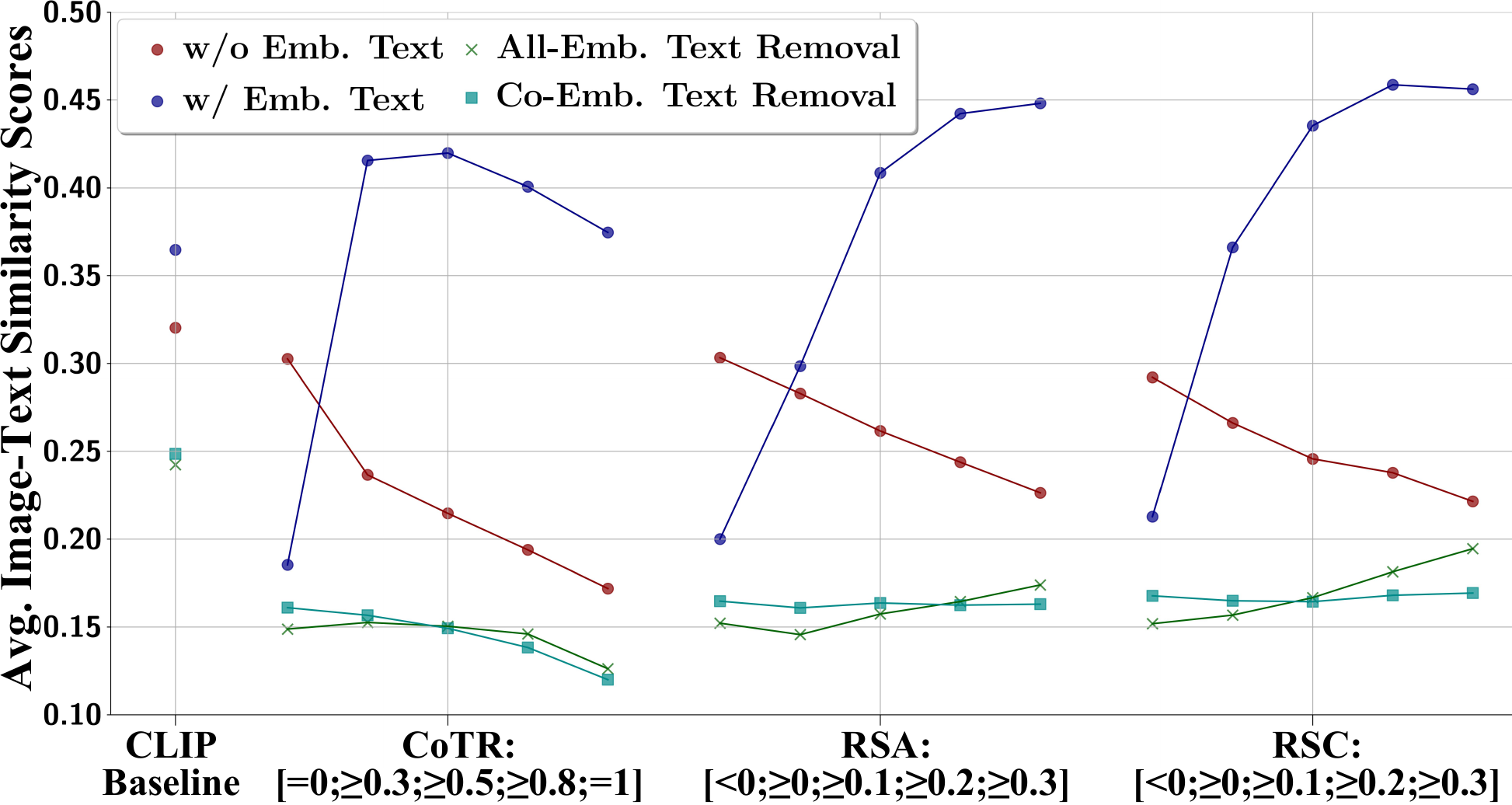}
\caption{\textbf{Comparison of mean similarity of LAION-2B subset for text spotting capacity validation.}
Models trained with more parrot captions are better at aligning the image with parrot captions but perform worse at aligning images without embedded text.
}
\label{fig:subset_real}
\end{figure}

\begin{table*}[]
\centering
\scalebox{0.8}{
\begin{tabular}{c|ccc|cc|cc|cc}
\toprule
 &
  \multicolumn{3}{c|}{\textbf{Visual Question Answering (Acc)}} &
  \multicolumn{2}{c|}{\textbf{Image Captioning (CIDEr)}} &
  \multicolumn{2}{c|}{\textbf{Text-to-Image Retrieval(R@1)}} &
  \multicolumn{2}{c}{\textbf{Image-to-Text Retrieval(R@1)}} \\
\multirow{-2}{*}{\begin{tabular}[c]{@{}c@{}}\textbf{BLIP}\\ \textbf{Data (3M)}\end{tabular}} &
  \textbf{VQAv2} &
  \cellcolor[HTML]{EFEFEF}{\color[HTML]{343434} \textbf{TextVQA}} &
  \cellcolor[HTML]{EFEFEF}{\color[HTML]{343434} \textbf{ST-VQA}} &
  \textbf{COCO} &
  \cellcolor[HTML]{EFEFEF}\textbf{TextCaps} &
  \textbf{COCO} &
  \cellcolor[HTML]{EFEFEF}\textbf{TextCaps} &
  \textbf{COCO} &
  \cellcolor[HTML]{EFEFEF}\textbf{TextCaps} \\ \hline
Rand &
  71.07 &
  \cellcolor[HTML]{EFEFEF}15.36 &
  \cellcolor[HTML]{EFEFEF}10.48 &
  115.6 &
  \cellcolor[HTML]{EFEFEF}53.7 &
  48.91 &
  \cellcolor[HTML]{EFEFEF}56.34 &
  65.46 &
  \cellcolor[HTML]{EFEFEF}72.45 \\
w/ Emb. Text &
  68.94 &
  \cellcolor[HTML]{EFEFEF}\textbf{19.05} &
  \cellcolor[HTML]{EFEFEF}\textbf{12.65} &
  108.9 &
  \cellcolor[HTML]{EFEFEF}\textbf{92.1} &
  42.89 &
  \cellcolor[HTML]{EFEFEF}\textbf{70.1} &
  58.5 &
  \cellcolor[HTML]{EFEFEF}\textbf{81.42} \\
w/o Emb. Text &
  \textbf{71.22} &
  \cellcolor[HTML]{EFEFEF}13.65 &
  \cellcolor[HTML]{EFEFEF}9.29 &
  \textbf{116.2} &
  \cellcolor[HTML]{EFEFEF}41.5 &
  \textbf{49.96} &
  \cellcolor[HTML]{EFEFEF}31.83 &
  \textbf{66.5} &
  \cellcolor[HTML]{EFEFEF}48.7 \\ \hline
CoTR = 0.0 &
  \textbf{71.11} &
  \cellcolor[HTML]{EFEFEF}13.97 &
  \cellcolor[HTML]{EFEFEF}9.75 &
  {\color[HTML]{333333} \textbf{116.3}} &
  \cellcolor[HTML]{EFEFEF}44.6 &
  \textbf{49.55} &
  \cellcolor[HTML]{EFEFEF}38.05 &
  \textbf{66.08} &
  \cellcolor[HTML]{EFEFEF}54.57 \\
CoTR $\geq$ 0.3 &
  67.4 &
  \cellcolor[HTML]{EFEFEF}19.28 &
  \cellcolor[HTML]{EFEFEF}11.81 &
  104.9 &
  \cellcolor[HTML]{EFEFEF}\textbf{96.9} &
  37.78 &
  \cellcolor[HTML]{EFEFEF}\textbf{67.28} &
  51.98 &
  \cellcolor[HTML]{EFEFEF}\textbf{78.2} \\
CoTR $\geq$ 0.5 &
  67.02 &
  \cellcolor[HTML]{EFEFEF}\textbf{19.64} &
  \cellcolor[HTML]{EFEFEF}12.38 &
  102.7 &
  \cellcolor[HTML]{EFEFEF}94.1 &
  35.94 &
  \cellcolor[HTML]{EFEFEF}65.24 &
  50.32 &
  \cellcolor[HTML]{EFEFEF}76.94 \\
CoTR $\geq$ 0.8 &
  66.38 &
  \cellcolor[HTML]{EFEFEF}18.50 &
  \cellcolor[HTML]{EFEFEF}12.00 &
  100.9 &
  \cellcolor[HTML]{EFEFEF}91.6 &
  34.13 &
  \cellcolor[HTML]{EFEFEF}62.65 &
  46.9 &
  \cellcolor[HTML]{EFEFEF}73.56 \\
CoTR = 1.0 &
  66.18 &
  \cellcolor[HTML]{EFEFEF}18.47 &
  \cellcolor[HTML]{EFEFEF}\textbf{12.80} &
  101.2 &
  \cellcolor[HTML]{EFEFEF}91.3 &
  33.55 &
  \cellcolor[HTML]{EFEFEF}61.83 &
  46.62 &
  \cellcolor[HTML]{EFEFEF}73.05 \\ \hline
RSA $<$ 0.0 &
  \textbf{70.79} &
  \cellcolor[HTML]{EFEFEF}14.16 &
  \cellcolor[HTML]{EFEFEF}9.64 &
  \textbf{115.7} &
  \cellcolor[HTML]{EFEFEF}44.9 &
  \textbf{48.25} &
  \cellcolor[HTML]{EFEFEF}36.85 &
  \textbf{64.72} &
  \cellcolor[HTML]{EFEFEF}54.7 \\
RSA $\geq$ 0.0 &
  70.03 &
  \cellcolor[HTML]{EFEFEF}18.76 &
  \cellcolor[HTML]{EFEFEF}11.81 &
  111.9 &
  \cellcolor[HTML]{EFEFEF}84.5 &
  46.25 &
  \cellcolor[HTML]{EFEFEF}\textbf{68.61} &
  62.92 &
  \cellcolor[HTML]{EFEFEF}\textbf{81.23} \\
RSA $\geq$ 0.1 &
  68.14 &
  \cellcolor[HTML]{EFEFEF}19.48 &
  \cellcolor[HTML]{EFEFEF}\textbf{13.33} &
  105.6 &
  \cellcolor[HTML]{EFEFEF}\textbf{96.1} &
  39.96 &
  \cellcolor[HTML]{EFEFEF}68.13 &
  54.64 &
  \cellcolor[HTML]{EFEFEF}79.37 \\
RSA $\geq$ 0.2 &
  66.01 &
  \cellcolor[HTML]{EFEFEF}\textbf{21.06} &
  \cellcolor[HTML]{EFEFEF}11.85 &
  98.7 &
  \cellcolor[HTML]{EFEFEF}94.4 &
  33.03 &
  \cellcolor[HTML]{EFEFEF}64.17 &
  47.12 &
  \cellcolor[HTML]{EFEFEF}75.33 \\
RSA $\geq$ 0.3 &
  64.20 &
  \cellcolor[HTML]{EFEFEF}18.44 &
  \cellcolor[HTML]{EFEFEF}12.04 &
  95.26 &
  \cellcolor[HTML]{EFEFEF}91.1 &
  26.64 &
  \cellcolor[HTML]{EFEFEF}60.11 &
  37.3 &
  \cellcolor[HTML]{EFEFEF}70.24 \\ \hline
RSC $<$ 0.0 &
  \textbf{70.13} &
  \cellcolor[HTML]{EFEFEF}15.19 &
  \cellcolor[HTML]{EFEFEF}10.74 &
  \textbf{112.2} &
  \cellcolor[HTML]{EFEFEF}46.7 &
  \textbf{46.8} &
  \cellcolor[HTML]{EFEFEF}41.95 &
  \textbf{63.24} &
  \cellcolor[HTML]{EFEFEF}58.05 \\
RSC $\geq$ 0.0 &
  68.86 &
  \cellcolor[HTML]{EFEFEF}20.12 &
  \cellcolor[HTML]{EFEFEF}\textbf{13.75} &
  107.8 &
  \cellcolor[HTML]{EFEFEF}93.5 &
  42.0 &
  \cellcolor[HTML]{EFEFEF}\textbf{69.78} &
  57.42 &
  \cellcolor[HTML]{EFEFEF}\textbf{80.92} \\
RSC $\geq$ 0.1 &
  67.35 &
  \cellcolor[HTML]{EFEFEF}\textbf{20.54} &
  \cellcolor[HTML]{EFEFEF}12.84 &
  103.4 &
  \cellcolor[HTML]{EFEFEF}\textbf{96.9} &
  36.4 &
  \cellcolor[HTML]{EFEFEF}66.69 &
  51.02 &
  \cellcolor[HTML]{EFEFEF}77.79 \\
RSC $\geq$ 0.2 &
  62.62 &
  \cellcolor[HTML]{EFEFEF}20.32 &
  \cellcolor[HTML]{EFEFEF}13.14 &
  98.7 &
  \cellcolor[HTML]{EFEFEF}92.8 &
  30.08 &
  \cellcolor[HTML]{EFEFEF}61.38 &
  42.96 &
  \cellcolor[HTML]{EFEFEF}71.98 \\
RSC $\geq$ 0.3 &
  63.75 &
  \cellcolor[HTML]{EFEFEF}18.94 &
  \cellcolor[HTML]{EFEFEF}13.03 &
  92.9 &
  \cellcolor[HTML]{EFEFEF}88.7 &
  24.23 &
  \cellcolor[HTML]{EFEFEF}58.35 &
  34.72 &
  \cellcolor[HTML]{EFEFEF}68.95 \\     \bottomrule
\end{tabular}}
\caption{\textbf{BLIP downstream tasks performance of pre-training on different curated 3M subsets.} The gray color represents tasks requiring the model to read the text from images.}
\label{tab:blip_exp}
\end{table*}

\noindent \textbf{Curation II: Co-Emb.~Text Rate (CoTR).}
Tab.~\ref{tab:ab_embtext} reports the CLIP models trained on parrot captions with different CoTR.
We first select all the images with embedded text and then sample images based on the CoTR depicted at Algorithm~\ref{alg:code} with different thresholds.
With increasing CoTR, all the zero-shot benchmark performance drops significantly.
Despite the images in the subset (CoTR = 0) all embedded with text, the pre-trained model performs similarly to the random baseline in~\ref{tab:ab_embtext}.
It indicates that the parrot caption is more crucial than embedded text in reducing the pre-trained data quality.
For the text spotting capacity, Fig.~\ref{fig:subset_syn} and~\ref{fig:subset_real} show that the increasing CoTR does not lead to stronger text spotting capacity, possibly due to the average length of captions decreasing in higher CoTR data.

\noindent \textbf{Curation III: Relative Score from Text Removal.}
Given the observations in Sec.~\ref{subsec:ab_remove}, we further select a series of subsets based on the relative score of images before and after text removal.
The subsets with higher relative scores are more dominant embedded text (RSA) or parrot captions (RSC) in CLIP score measurement.
In Tab.~\ref{tab:ab_rsa} and~\ref{tab:ab_rsc}, we report the zero-shot performance of models trained on subsets with different relative score thresholds.
The CLIP pre-trained with higher RSA or RSC both get worse downstream performance.
Importantly, the average raw CLIP scores S({\color{raw_mark}{\large$\bullet$}}) of these subsets have a positive correlation with RSA or RSC, indicating using CLIP scores from a biased pre-trained model as the data filtering strategy can be unreliable.
When comparing the RSA and RSC, the results show that the samples dominated by the latter, i.e., parrot captions, are less informative for CLIP training.
Moreover, Fig.~\ref{fig:subset_syn} and.~\ref{fig:subset_real} show that the text spotting capacity of CLIP can be further improved by training on the samples using relative scores as data curation criteria against CoTR.

\subsection{Ablation Study on Text-Oriented Tasks}
Inspired by~\cite{ganz2023towards}, we further investigate the model behavior on downstream tasks requiring reading text, including Visual Question Answering (VQA), Image Captioning, and Text-Image Retrieval.
Specifically, for the text-oriented tasks, we use Text VQA~\cite{singh2019towards} and ST-VQA~\cite{biten2019scene} for VQA, and TextCaps~\cite{sidorov2020textcaps} for captioning and retrieval.
Moreover, we also provide the same tasks on the datasets that only require the model to see, i.e., the natural image dataset.
Similarly, we use VQAv2~\cite{goyal2017making} for VQA and COCO~\cite{chen2015microsoft} for captioning and retrieval.
We choose BLIP~\cite{li2022blip} for the ablation study instead of CLIP as it can be directly applied to all these tasks.
We first pre-train the BLIP on different subsets with 10 epochs and then finetune 10 epochs for VQA, and 5 epochs for captioning and retrieval.
As shown in Tab.\ref{tab:blip_exp}, training BLIPs to spot text can boost their performance on the downstream tasks requiring the model to read but impede the performance of downstream tasks only requiring the model to see, which are consistent with the observation on classification tasks.
Nevertheless, when BLIPs mainly focus on reading, e.g. (RSA $\geq$ 0.3), their text-oriented and natural downstream performance also decreases.
In other words, the parrot captions can benefit the text-orient downstream tasks while requiring careful data mixing trade-off.

\section{Profiling More Image-Text Dataset}

\noindent \textbf{MMC4.} Multimodal C4 (MMC4)~\cite{zhu2023multimodal} is a popular image-text interleaved dataset also built on the CLIP feature matching.
A linear assignment algorithm is used to place images into longer bodies of text using CLIP features.
Therefore, we profile the MMC4 dataset with the same pipeline in Sec.~\ref{sec:data} to investigate whether parrot captions commonly exist.
Note that, we calculate the assigned text for Co-Emb.~text statistics.
As shown in Tab.~\ref{table:mmc4_prostat}, we found that the image distribution in MMC4 is similar to LAION-2B, with around 50\% of images containing embedded text.
Meanwhile, the average captions of MMC4 are much longer than those of LAION-2B, resulting in a lower CoTR than LAION-2B.
Nevertheless, how the correlation between embedded text and images affects the interleaved dataset still needs further investigation, which we left for future work.

\noindent \textbf{CC12M.}
Conceptual 12M (CC12M)~\cite{changpinyo2021conceptual} dataset is built on a rule-based system without using CLIP models from webpages
annotations.
We further profile the CC12M to ablate the origin of parrot captions in different data curation pipelines.
Tab.~\ref{table:cc_prostat} shows that the text is commonly embedded in the web images, while the Co-Emb.~text rate is more lower than the LAION-2B~\cite{schuhmann2022laion} dataset.
Therefore, there is still room for improving the data collection pipelines based on the CLIP model filtering.

\begin{table}[t]
\centering
\begin{tabular}{l|c}
\toprule
Number of Total Images & 527156206 \\
Number of Images w/ Emb.~Text & 264618807 \\
Image w/ Emb.~Text Rate & 50.20\% \\
\hline
Co-Emb.~Text Rate (in Total) & 2.88\% \\
-- (in Images w/ Emb.~Text) & 15.70\% \\
Fuzzy Co-Emb.~Text Rate (in Total) & 5.75\% \\
-- (in Images w/ Emb.~Text) & 31.28\%  \\
\bottomrule
\end{tabular}

\caption{
\textbf{Overall parrot captions statistic in MMC4~\cite{zhu2023multimodal}}.
}
\label{table:mmc4_prostat}
\end{table}
\begin{table}[t]
\centering
\begin{tabular}{l|c}
\toprule
Number of Total Images & 9,230,079 \\
Number of Images w/ Emb.~Text & 3,421,152 \\
Image w/ Emb.~Text Rate & 37.06\% \\
\hline
Co-Emb.~Text Rate (in Total) & 6.21\% \\
-- (in Images w/ Emb.~Text) & 15.94\% \\
Fuzzy Co-Emb.~Text Rate (in Total) & 16.75\% \\
-- (in Images w/ Emb.~Text) & 43.01\%  \\
\bottomrule
\end{tabular}

\caption{
\textbf{Overall parrot captions statistic in CC12M~\cite{changpinyo2021conceptual}}.
}
\label{table:cc_prostat}
\end{table}

\renewcommand{\thefootnote}{\arabic{footnote}}
\begin{table*}[tp]
    \centering 
\begin{tabular}{c|c|cc|cc}
\toprule

\textbf{Metric} & \textbf{Ours} & \textbf{CLIP} & \textbf{OpenCLIP} & \textbf{DC medium} & \textbf{DC large}\\\hline
Data & 100M (LAION) & 400M (WIT) & 2B (LAION) & 128M (DC) & 1.28B (DC)\\ \hline
Sync. Score $\downarrow$ & 0.163 $\pm$ 0.065 & 0.317 $\pm$ 0.030 & 0.368 $\pm$ 0.042& 0.268 $\pm$ 0.024& 0.338 $\pm$ 0.034\\
\hline
IN & 0.526 & 0.633 & 0.666 & 0.176 & 0.459\\
IN dist. shifts & 0.404 & 0.485 & 0.522 & 0.152 & 0.378\\
VTAB & 0.481 & 0.526 & 0.565 & 0.259 & 0.426\\
Retrieval & 0.421 & 0.501 & 0.560 & 0.219 & 0.419\\
Avg.~38 datasets & 0.443 & 0.525 & 0.565 & 0.258 & 0.437
\\
\bottomrule
\end{tabular}
    \caption{\textbf{Comparison of our debiased model and the released pre-trained models.} We evaluate on our proposed synthetic (Sec.~\ref{sec:ab_syn}) and Datacomp~\cite{gadre2023datacomp} benchmark. For the synthetic benchmark, we use the 1-gram vocabulary built from captions and report the mean and std of the synthetic image-text similarity (Sync. S). We also report the performance of CLIP model trained on medium and large Datacomp~\cite{gadre2023datacomp}(DC) pools with no filtering.}
    \label{tab:100m_clip}
\end{table*}
\section{A Simple Fix with Text-Orientated Filtering}

To provide an alternative solution for existing released CLIP models, we further construct a less text-biased LAION-2B subset by filtering the dataset with OCR results.
Specifically, we first remove all images with detected text.
Then, we filter the image-text pairs with a CLIP score greater than 0.3 and an aesthetics score greater than 0.45 to obtain high-quality data.
Finally, we perform deduplication in each cluster based on the K-means label predicted in Sec.~\ref{sec:data} to obtain the compact filtering dataset with 107,166,507 (100M) samples.
Given the curated subset, we train a CLIP model from scratch following the setting used in~\cite{schuhmann2022laion}.
The performance of our trained CLIP is reported in Tab.~\ref{tab:100m_clip}.
It indicates that the CLIP model can achieve a high performance while without introducing such text spotting bias.
Nevertheless, due to the imperfect OCR results, the subset inevitably contains some parrot captions and brings costly scalability, which we also left for future work. The pre-trained models~\footnote{\tt \scriptsize \url{https://github.com/opendatalab/CLIP-Parrot-Bias}} and the filtered subset~\footnote{\tt \scriptsize \url{https://openxlab.org.cn/datasets/opendatalab-linyiqi/LAION-text-debiased-100M}} are released on OpenDataLab~\cite{conghui2022opendatalab}.
\begin{figure}[tp]
    \centering
    \includegraphics[width=\linewidth]{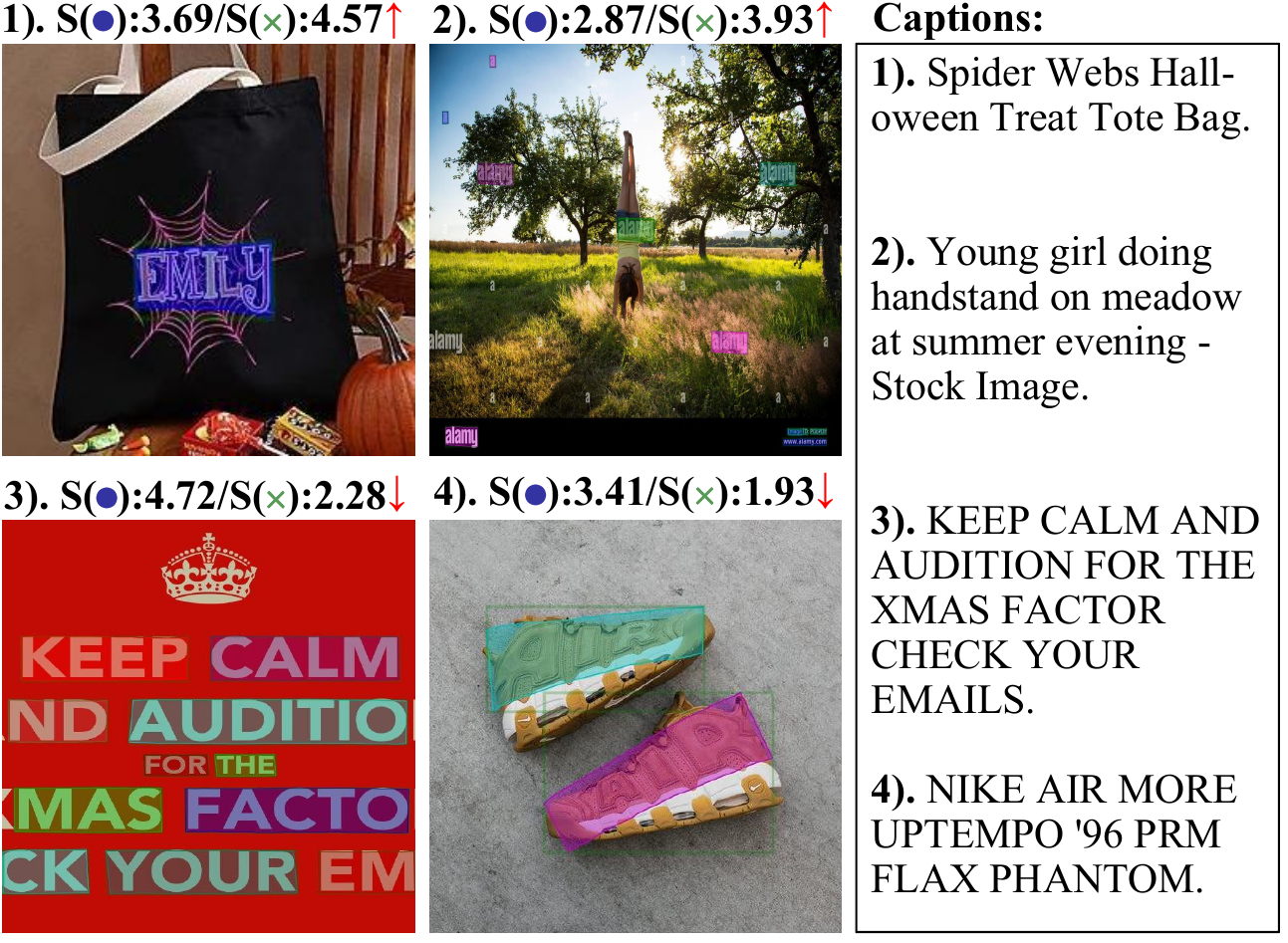}
    \caption{\textbf{Embedded texts play different roles for visual concepts.}
    S({\color{raw_mark}{\large$\bullet$}}) and S({\color{all_mark}\textbf{{$\times$}}}) denote the CLIP score before and after removing All-Emb.~text.
    OCR results are masked with color.
    }
    \label{fig:case_std}
\end{figure}

\section{Discussion and Conclusion}
The popularity of vision-language contrastive loss stems from its efficiency and simplicity.
However, the analysis and experiments we presented show that the embedded text in images and their parrot captions plant significant text-spotting bias due to such contrastive fashions.
Firstly, almost half of the captions in the widely used LAION-2B dataset are biased towards parroting the embedded text in images.
Secondly, the pre-trained CLIP models have strong preferences for the image-text pair with parrot captions, which achieve higher similarity scores than those without.
Finally, using data biasing to parrot captions, we can easily train a CLIP model with a strong text-spotting bias.
Our work demonstrates the emergency of reviewing the impact of parrot captions in the entire ecosystem of CLIP models.

Here, we further showcase some examples to provide a more complete perspective on their negative impact and functionalities.
On the one hand, as shown in Fig.~\ref{fig:case_std}, when the embedded text is not directly relevant to visual content, like `EMILY' on the bag and watermark, this text plays a strong distractor for the CLIP alignments.
On the other hand, parts of web media content and the concept propagation inherently are presented by embedded text, such as slogans and brand logos in Fig.~\ref{fig:case_std}.
Therefore, our future endeavor involves building a bias-aware data curation pipeline and a more robust training function to mitigate such issues. Again, we urge the community to acknowledge and address these issues with greater attention.

{
    \small
    \bibliographystyle{ieeenat_fullname}
    \bibliography{main}
}
\clearpage
\setcounter{page}{1}
\maketitlesupplementary

\section{More Experiments on CC12M Dataset}

We further examine the CC12M~\cite{changpinyo2021conceptual} dataset with the same process in Sec.~\ref{sec:pretrained} and ~\ref{sec:training} to investigate whether parrot captions dominate the CLIP models' behavior in different data curation pipelines.
The overall CLIP scores statistics and experiments are shown in Tab.~\ref{table:cc_stat} and Tab.~\ref{tab:3m_cc}.
The results show that the captions in CC12M are less correlated with text embedded in the images based on the mean CLIP scores.
The CLIP training results on different curated subsets indicate that the embedded text in CC12M only slightly reduces the pre-training dataset quality, while the images with embedded text still harm the downstream performance. 
In summary, the CC12M dataset constructed without CLIP score is not significantly biased towards parrot captions.

\begin{table}[t]
\centering
\begin{tabular}{l|c}
\toprule
Mean CLIP scores (Whole Dataset) & 0.3021\\
Mean CLIP scores (All-Emb.~Text Removal) & 0.2912\\
Mean CLIP scores (Co-Emb.~Text Removal) & 0.2892\\
\bottomrule
\end{tabular}

\caption{The mean CLIP scores of CC12M, which are obtained from ViT-B-32 models. The text removal operations are the same as Sec.~\ref{sec:pretrained}, while the results are from the whole dataset.
}
\label{table:cc_stat}
\end{table}

\begin{table}[tp]
    \centering 
\scalebox{0.9}{
    \begin{tabular}{l|c|ccc|c}
    \toprule
    Data (3M) & Model &  IN & Ret. & Avg. & Sync. S\\ \hline
    Random & RN50 & 0.205 & \cellcolor{tabfirst}0.253& 0.229 & 0.186\\
    w/o Emb.~Text & RN50 & \cellcolor{tabfirst}0.206 & 0.248 & \cellcolor{tabfirst}0.231 & 0.121\\
    w/ Emb.~Text Only & RN50 & 0.161 & 0.232 & 0.210 & \cellcolor{tabfirst}0.220\\
    \hline
    Random & ViT-B   & 0.142 & \cellcolor{tabfirst}0.193 & 0.206 & 0.127\\
    w/o Emb.~Text & ViT-B  &\cellcolor{tabfirst}0.151 & 0.190 & \cellcolor{tabfirst}0.214 & 0.096\\
    w/ Emb.~Text Only & ViT-B  & 0.113 & 0.165 & 0.196 & \cellcolor{tabfirst}0.148\\
    \bottomrule
    \end{tabular}}
    \caption{\textbf{Comparison of dataset quality on sampled subsets.}
    The subsets are sampled the same as Sec.~\ref{sec:ab_std} Curation I.
    The Sync.S denotes the average CLIP score of syn-emb.~text benchmark in Sec.~\ref{sec:ab_std}.
    }
    \label{tab:3m_cc}
\end{table}

\begin{table}[tp]
    \centering 
\scalebox{1}{
    \begin{tabular}{l|c|ccc}
    \toprule
    Data (12M) & Model &  IN & Ret. & Avg. \\ \hline
    CoTR = 0.0 & RN50       & \cellcolor{tabfirst}0.349 & \cellcolor{tabfirst}0.367 & \cellcolor{tabfirst}0.348 \\
    CoTR  $\geq$ 0.3 & RN50      & 0.055 & 0.115 & 0.159 \\
    CoTR  $\geq$ 0.5 & RN50      & 0.037 & 0.102 & 0.135 \\
    CoTR  $\geq$ 0.8 & RN50      & 0.019 & 0.084 & 0.102 \\
    CoTR  = 1.0 & RN50     & 0.017 & 0.080 & 0.112 \\
    \hline
    CoTR  = 0.0 & ViT-B   & \cellcolor{tabfirst}0.302 & \cellcolor{tabfirst}0.303 & \cellcolor{tabfirst}0.320 \\
    CoTR  $\geq$ 0.3 & ViT-B  & 0.059 & 0.104 & 0.165 \\
    CoTR  $\geq$ 0.5 & ViT-B  & 0.040 & 0.098 & 0.141 \\
    CoTR  $\geq$ 0.8 & ViT-B  & 0.021 & 0.078 & 0.117 \\
    CoTR  = 1.0 & ViT-B & 0.021 & 0.081 & 0.114 \\
    \bottomrule
    \end{tabular}}
    \caption{\textbf{Ablation of different Co-Emb.~Text Rate(CoTR).}
    The fewer parrot captions, the better downstream task performance.
    }
    \label{tab:ab_cotr12m}
\end{table}

\begin{table}[tp]
\scalebox{0.9}{
    \centering
    \begin{tabular}{l|c|c|ccc}
    \toprule
    Data (3M) & Model  & Avg.S({\color{raw_mark}{\large$\bullet$}})&  IN & Ret. & Avg. \\ \hline
    RSA $<$ 0.0& RN50  &0.319 & \cellcolor{tabfirst}0.327 & \cellcolor{tabfirst}0.349 & \cellcolor{tabfirst}0.336 \\
    RSA $\geq$ 0.0& RN50 &0.339 & 0.245 & 0.294 & 0.292 \\
    RSA $\geq$ 0.1& RN50 &0.351 & 0.078 & 0.159 & 0.179 \\
    RSA $\geq$ 0.2& RN50 &0.360 & 0.028 & 0.101 & 0.125 \\
    RSA $\geq$ 0.3& RN50 &0.376 & 0.016 & 0.083 & 0.109 \\
    \hline
    RSA $<$ 0.0& ViT-B &0.319 & \cellcolor{tabfirst}0.277 & \cellcolor{tabfirst}0.285 & \cellcolor{tabfirst}0.313 \\
    RSA $\geq$ 0.0& ViT-B &0.339 & 0.211 & 0.241 & 0.285 \\
    RSA $\geq$ 0.1& ViT-B &0.351 & 0.068 & 0.133 & 0.180 \\
    RSA $\geq$ 0.2& ViT-B &0.360 & 0.024 & 0.090 & 0.120 \\
    RSA $\geq$ 0.3& ViT-B &0.376 & 0.011 & 0.076 & 0.103 \\
    \bottomrule
    \end{tabular}
    }
    \caption{\textbf{Ablation of models trained on subsets sampled by different RSA}. RSA denotes the relative similarity (S({\color{raw_mark}{\large$\bullet$}}) $\mathbf{-}$ S({\color{all_mark}\textbf{{$\times$}}})) of raw S({\color{raw_mark}{\large$\bullet$}}) and removed All-Emb.~text S({\color{all_mark}\textbf{{$\times$}}}) images.
    }
    \label{tab:ab_rsa12m}
\end{table}

\begin{table}[tp]
\scalebox{0.9}{
    \centering
    \begin{tabular}{l|c|c|ccc}
    \toprule
    Data (3M) & Model & Avg.S({\color{raw_mark}{\large$\bullet$}}) & IN & Ret. & Avg. \\ \hline 
    RSC $<$ 0.0& RN50 &0.326 & \cellcolor{tabfirst}0.125 & \cellcolor{tabfirst}0.171 & \cellcolor{tabfirst}0.209 \\
    RSC $\geq$ 0.0& RN50 &0.345 & 0.062 & 0.129 & 0.168 \\
    RSC $\geq$ 0.1&RN50 & 0.354 & 0.014 & 0.091 & 0.106 \\
    RSC $\geq$ 0.2& RN50 &0.364 & 0.008 & 0.084 & 0.104 \\
    RSC $\geq$ 0.3& RN50 &0.380 & 0.005 & 0.058 & 0.084 \\
    \hline
    RSC $<$ 0.0& ViT-B &0.326 & \cellcolor{tabfirst}0.079 & \cellcolor{tabfirst}0.129 & \cellcolor{tabfirst}0.174 \\
    RSC $\geq$ 0.0& ViT-B &0.345 & 0.045 & 0.119 & 0.149 \\
    RSC $\geq$ 0.1& ViT-B &0.354 & 0.018 & 0.091 & 0.116 \\
    RSC $\geq$ 0.2& ViT-B &0.364 & 0.008 & 0.076 & 0.106 \\
    RSC $\geq$ 0.3& ViT-B &0.380 & 0.004 & 0.059 & 0.091\\
    \bottomrule
    \end{tabular}
    }
    \caption{\textbf{Ablation of models trained on subsets sampled by different RSC.}
    RSC denotes the relative similarity (S({\color{raw_mark}{\large$\bullet$}}) $\mathbf{-}$ S({\color{co_mark}{\scriptsize$\blacksquare$}})) of raw S({\color{raw_mark}{\large$\bullet$}}) and removed Co-Emb.~text S({\color{co_mark}{\scriptsize$\blacksquare$}}) images.
    }
    
    \label{tab:ab_rsc12m}
\end{table}

\begin{figure*}[tp]
    \centering
    \includegraphics[width=\textwidth]{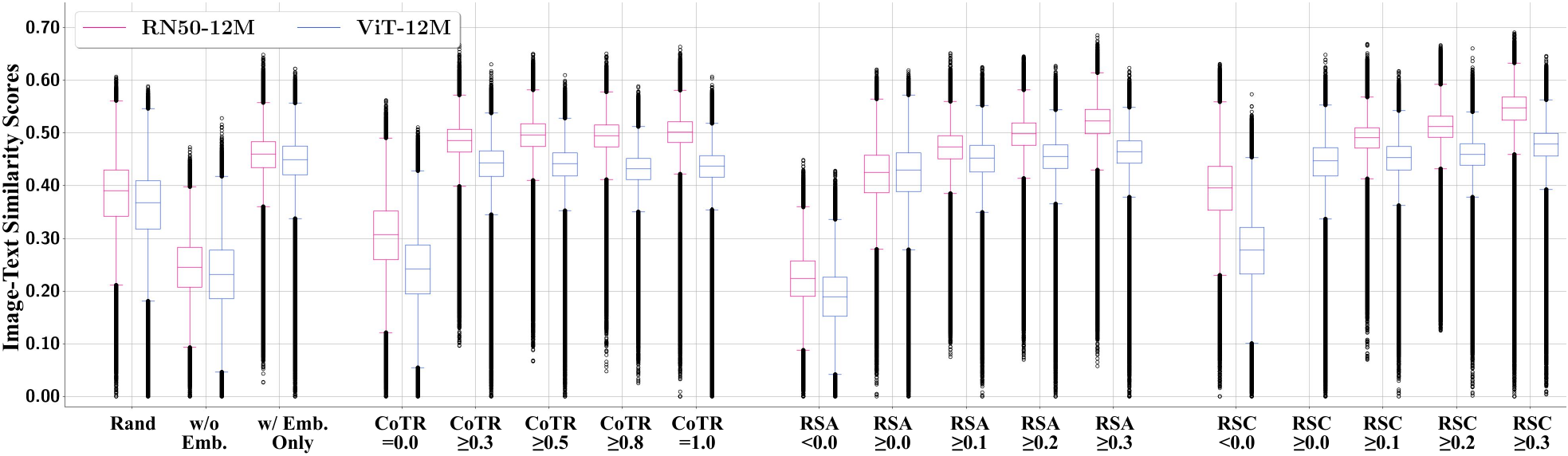}
    \caption{Benchmarking text spotting capacity of CLIP models on 12M scales with 1-gram caption vocabulary synthetic images dataset.}
    \label{fig:12M_text}
\end{figure*}

\begin{figure*}[tp]
\centering
\includegraphics[width=\textwidth]{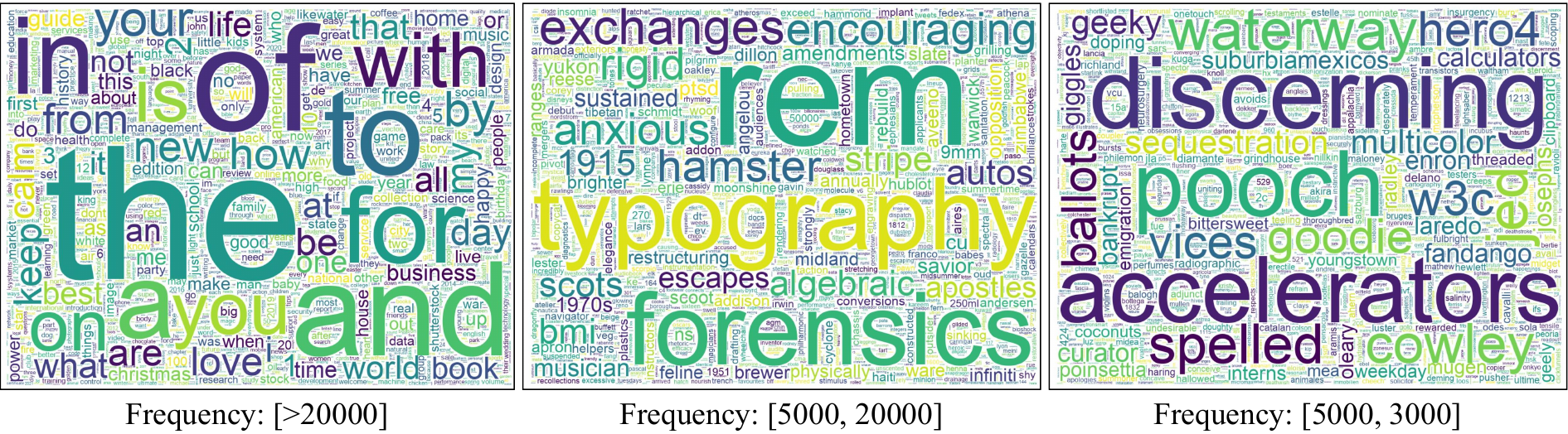}
\caption{The word clouds of 1-gram caption vocabulary.}
\label{fig:word_cloud}
\end{figure*}

\section{Technical Details}
For the PCA used in feature processing, we use Incremental PCA implemented by scikit-learn~\cite{scikit-learn}.
For the K-Means model, we train for 300 iterations and redo 10 times to select the best model.
For the CLIP model training, we used PyTorch DDP and amp precision to train models on a single machine with 8 NVIDIA A100 GPUs. 
We used AdamW~\cite{loshchilov2017decoupled} as an optimizer, with ${\beta}_1 = 0.9$ and ${\beta}_2 = 0.98$ for all models.
We used a cosine decay schedule with a linear warmup.
We used a resolution of 224 $\times$224 images for pre-training.
The training loss is the InfoNCE loss~\cite{radford2021learning}.

\section{Frequent Words in N-gram Vocabulary}
Fig.~\ref{fig:word_cloud} shows the word clouds of 1-gram caption vocabulary.
To better visualize the long-tail distribution, the word clouds are drawn from three different frequency intervals.
It shows that the 1-gram text becomes more complex and longer when the frequency is lower.

\begin{figure*}[tp]
    \centering
    \begin{subfigure}{\textwidth}
        \centering
    \includegraphics[width=\textwidth]{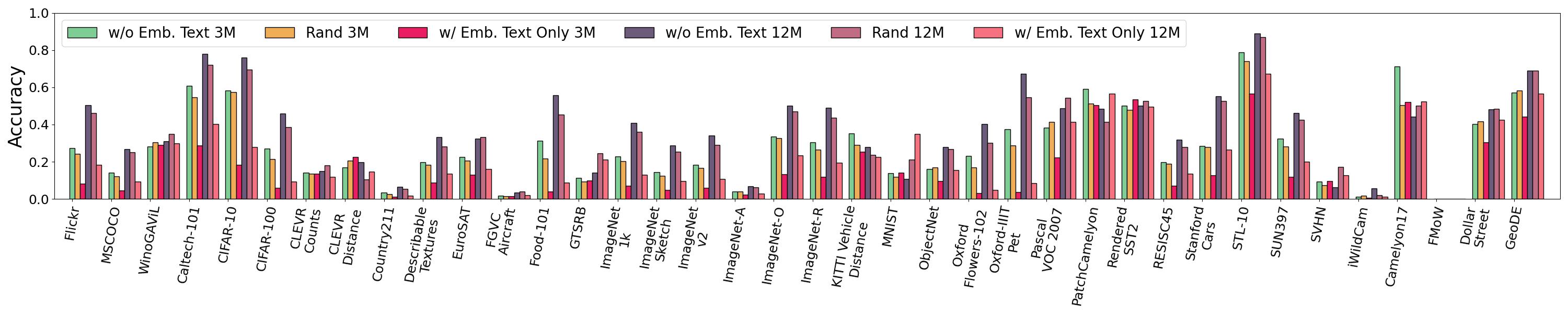}
    \caption{RN50}
    \end{subfigure}
    \hfill
    \begin{subfigure}{\textwidth}
        \centering
    \includegraphics[width=\textwidth]{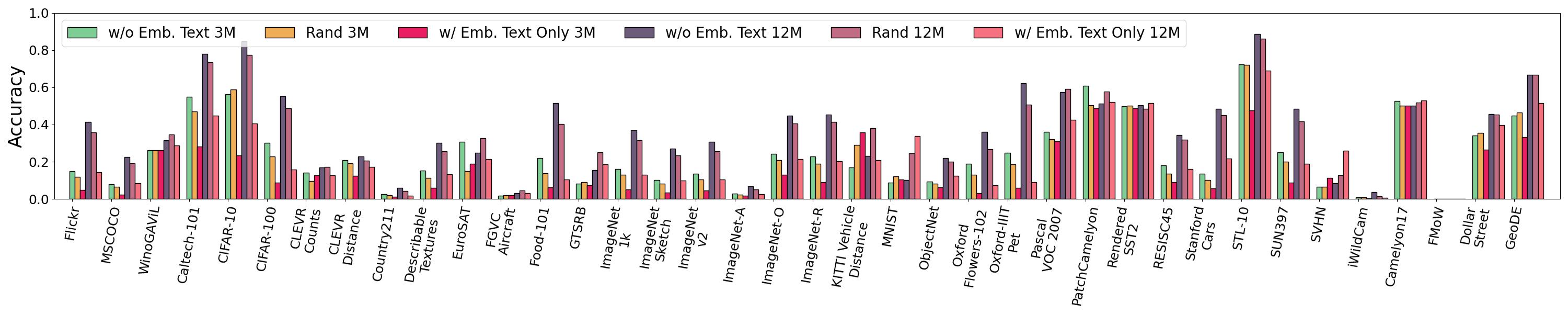}
    \caption{ViT-B-32}
    \end{subfigure}
    \caption{Full tracks DataComp evaluation of Curation I: Embedded Text in Images.}
    \label{fig:randexp}
\end{figure*}

\begin{figure*}[tp]
    \centering
    \begin{subfigure}{\textwidth}
    \centering
    \includegraphics[width=\textwidth]{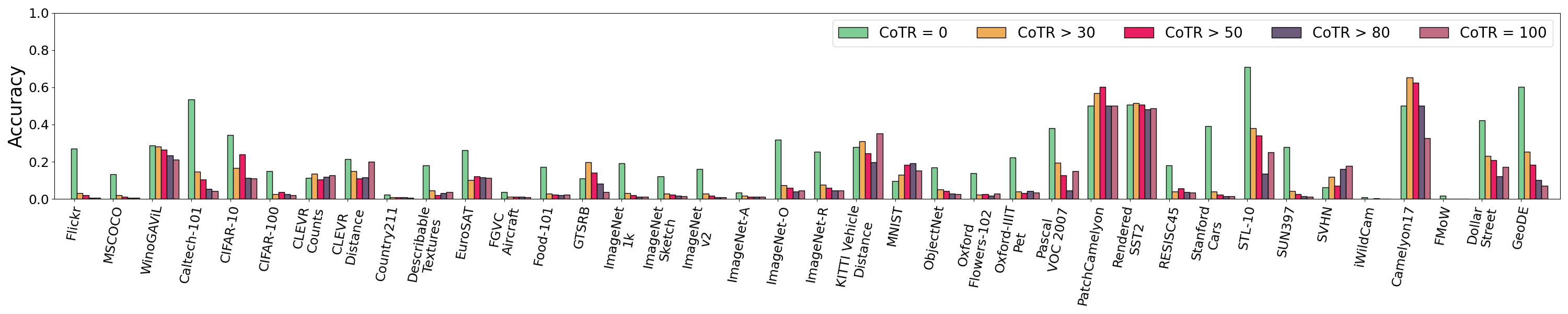}
    \caption{RN50}
    \end{subfigure}
    \hfill
    \begin{subfigure}{\textwidth}
    \centering
    \includegraphics[width=\textwidth]{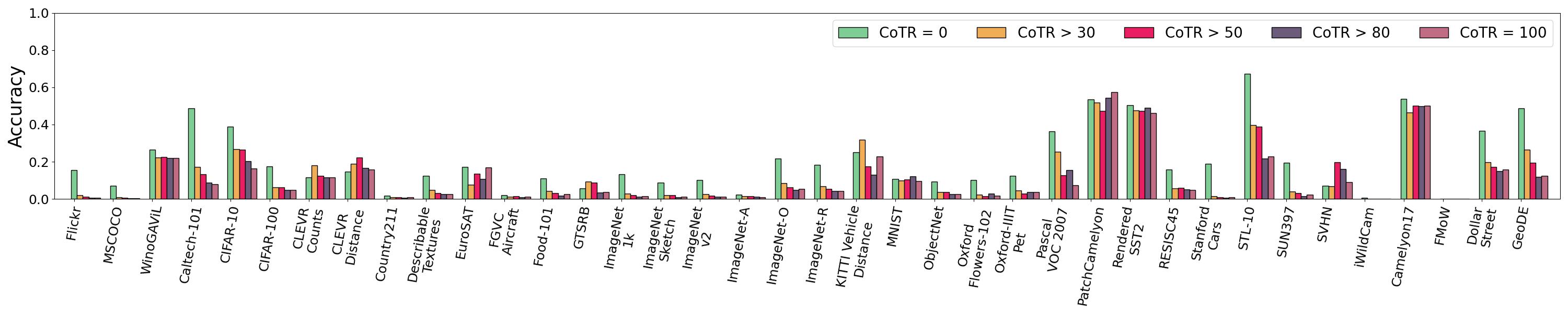}
    \caption{ViT-B-32}
    \end{subfigure}
    \caption{Full tracks DataComp evaluation of Curation II: Co-Emb. Text Rate (CoTR).}
    \label{fig:cotrexp}
\end{figure*}

\begin{figure*}[tp]
    \centering
    \begin{subfigure}{\textwidth}
        \centering
    \includegraphics[width=\textwidth]{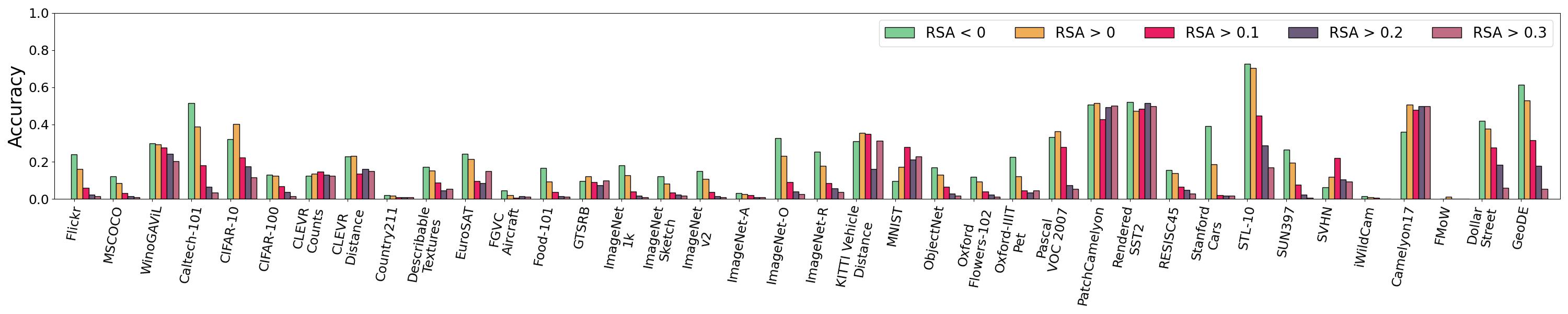}
        \caption{RN50 of RSA}
    \end{subfigure}
    \hfill
    \begin{subfigure}{\textwidth}
        \centering
        \includegraphics[width=\textwidth]{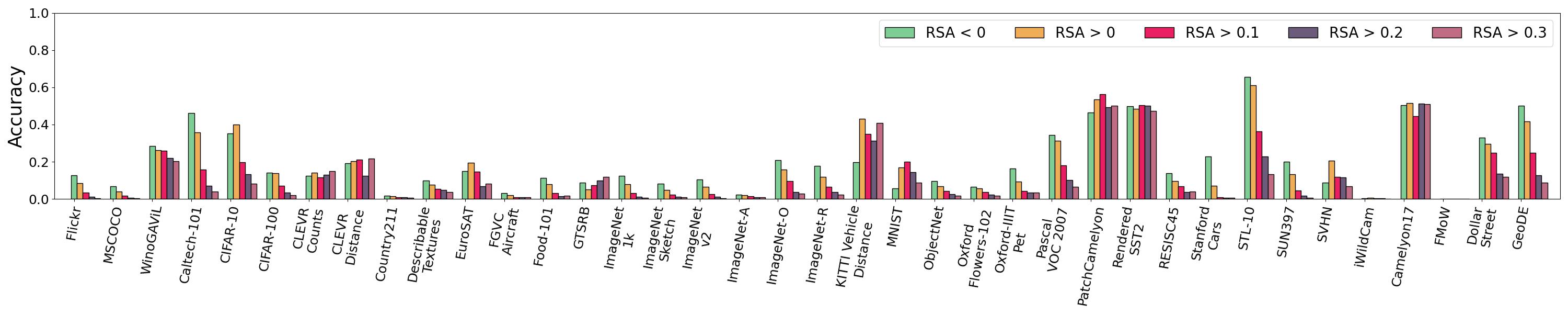}
        \caption{ViT-B-32 of RSA}
    \end{subfigure}%
    \hfill
    \begin{subfigure}{\textwidth}
        \centering
        \includegraphics[width=\textwidth]{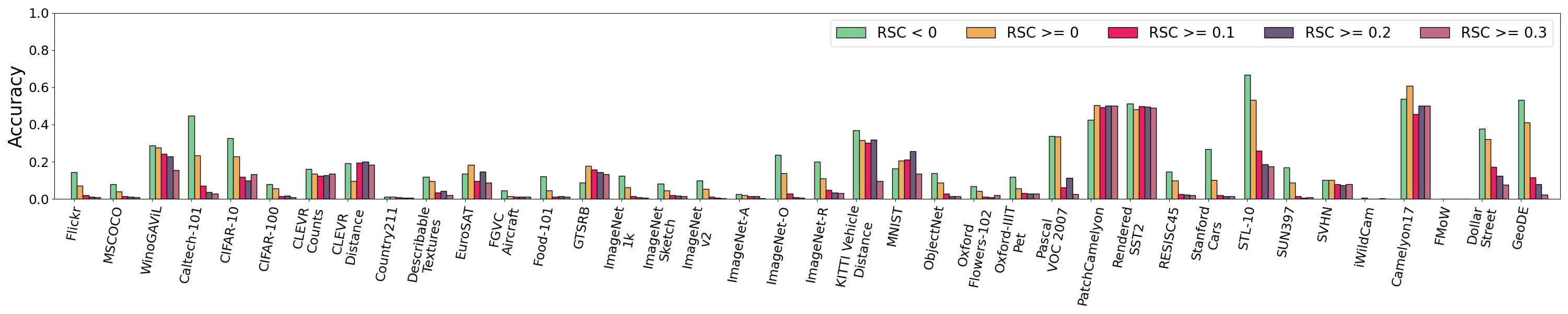}
        \caption{RN50 of RSC}
    \end{subfigure}%
    \hfill
    \begin{subfigure}{\textwidth}
        \centering
        \includegraphics[width=\textwidth]{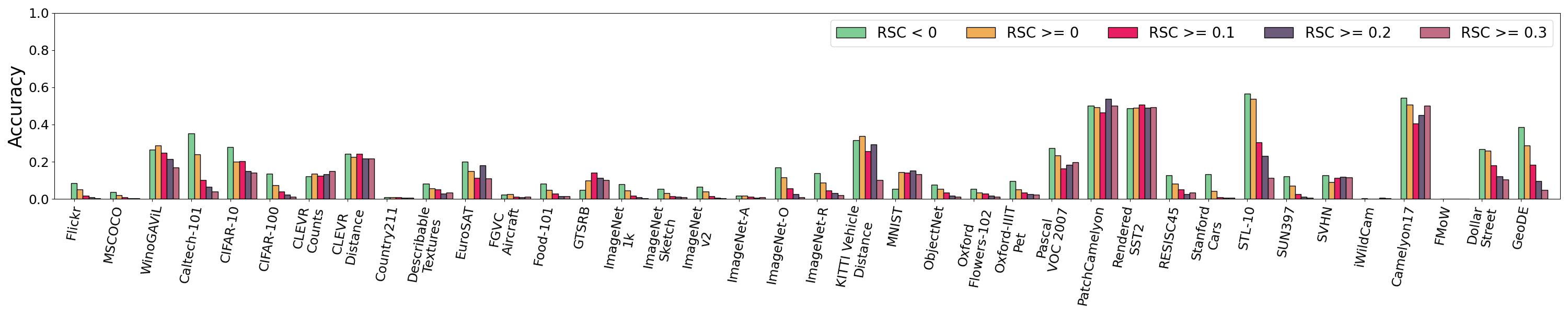}
        \caption{ViT-B-32 of RSC}
    \end{subfigure}
    \caption{Full tracks DataComp evaluation of Curation III: Relative Score from Text Removal.}
    \label{fig:relative}
\end{figure*}

\section{Curation Ablation on 12M scales}
We further provide the ablation study on 12M scales in Tab.~\ref{tab:ab_cotr12m},~\ref{tab:ab_rsa12m} and~\ref{tab:ab_rsc12m}.
All the results are consistent with the 3M scale results.
Fig.~\ref{fig:12M_text} reports the text spotting capacity of models on 12M scales using the synthetic benchmark the same as Sec.~\ref{sec:ab_std}.
It shows that training with more parrot caption samples does not lead to a stronger text spotting performance in synthetic benchmarks.

\section{Full Tracks Evaluation on DataComp}
In Fig.~\ref{fig:randexp}, ~\ref{fig:cotrexp}, and ~\ref{fig:relative} we report all dataset results on DataComp~\cite{gadre2023datacomp} of the ablation study.
In most vision-centric datasets, the model performance is consistent with the average performance.
Meanwhile, the results also indicate that the model with stronger text capacity achieves better performance on the text-oriented task, such as MINST.

\section{Sample Visualization}
\begin{figure*}[tp]
    \centering
    \begin{subfigure}{\textwidth}
        \centering
        \includegraphics[width=\textwidth]{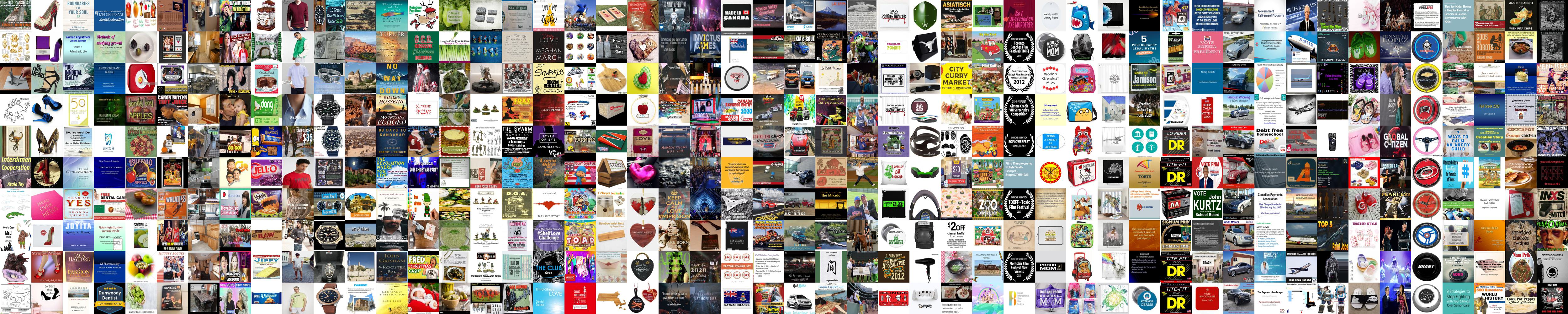}
        \caption{Cluster 0 to 50}
    \end{subfigure}
    \hfill
    \begin{subfigure}{\textwidth}
        \centering
        \includegraphics[width=\textwidth]{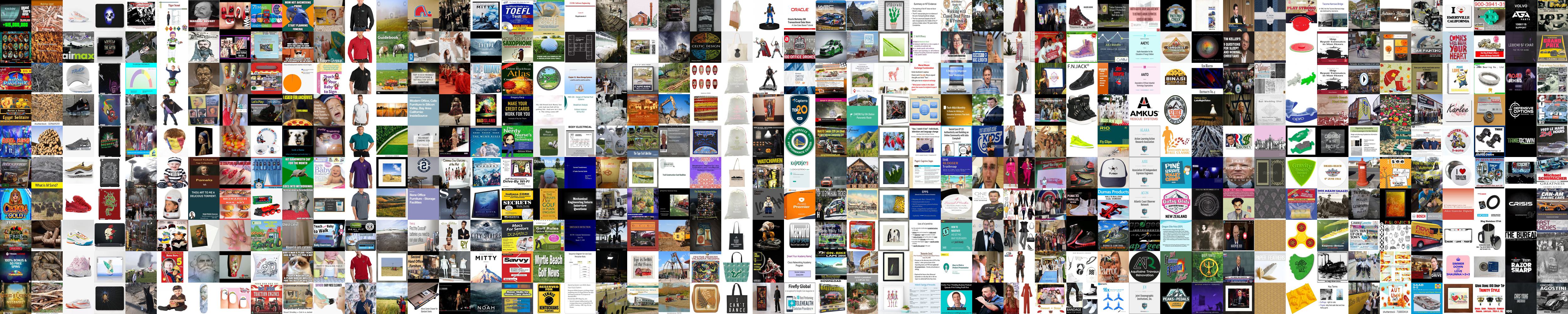}
        \caption{Cluster 50 to 100}
    \end{subfigure}%
    \hfill
    \begin{subfigure}{\textwidth}
        \centering
        \includegraphics[width=\textwidth]{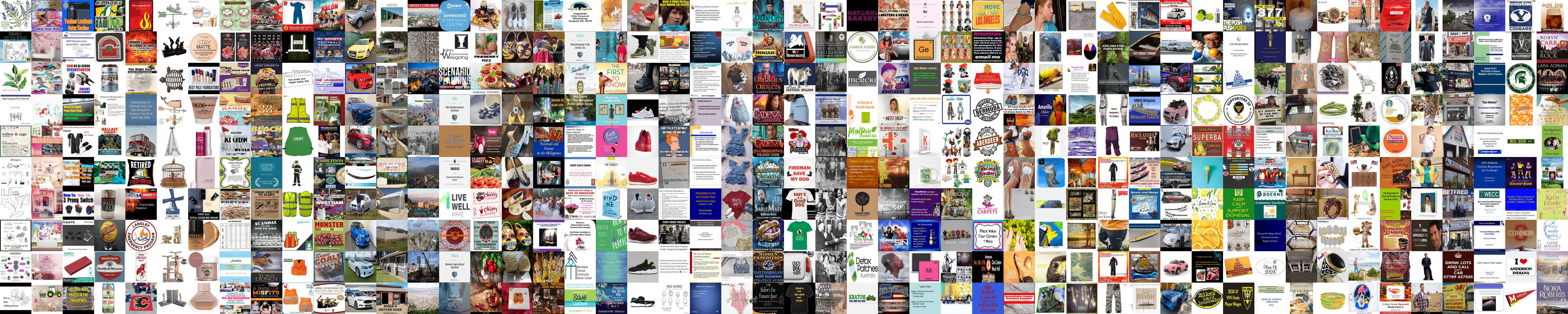}
        \caption{Cluster 100 to 150}
    \end{subfigure}%
    \hfill
    \begin{subfigure}{\textwidth}
        \centering
        \includegraphics[width=\textwidth]{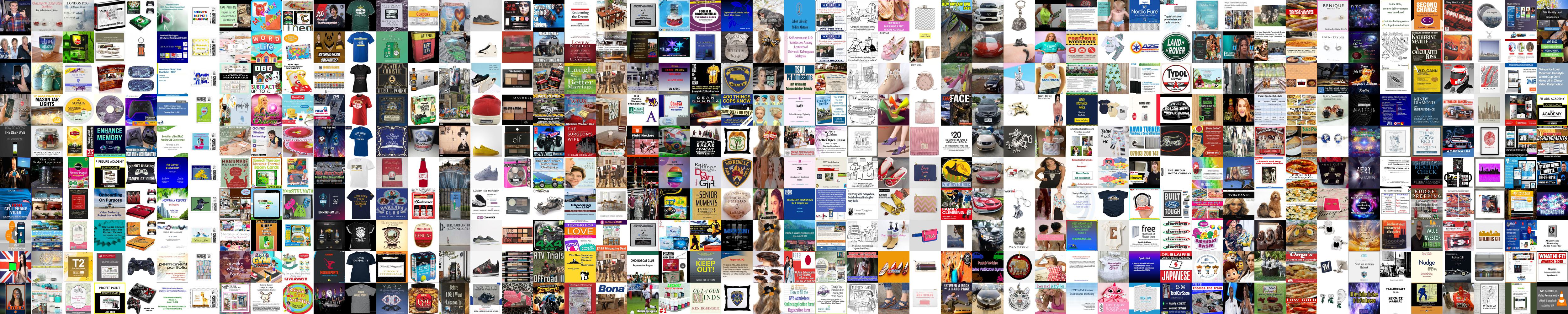}
        \caption{Cluster 150 to 200}
    \end{subfigure}
    \hfill
    \begin{subfigure}{\textwidth}
        \centering
        \includegraphics[width=\textwidth]{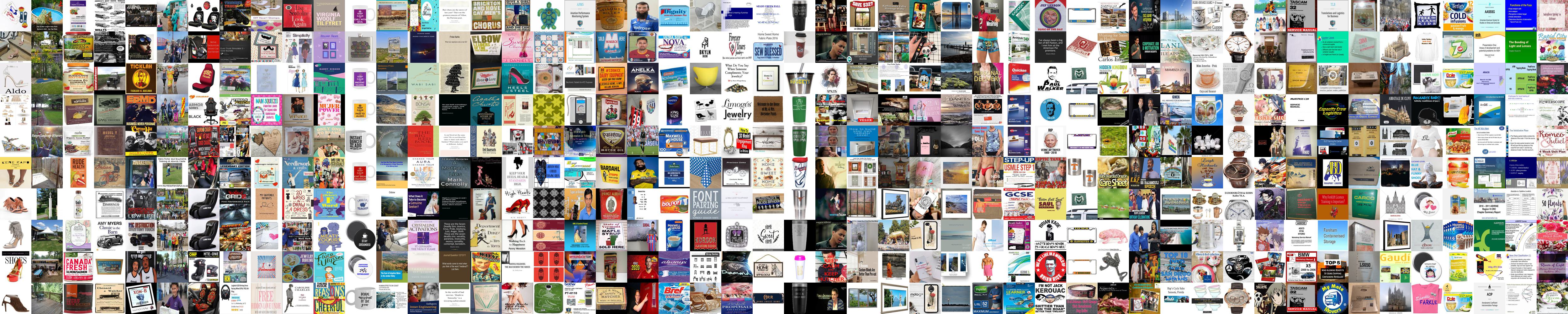}
        \caption{Cluster 200 to 250}
    \end{subfigure}
    \caption{\textbf{Top CLIP scores sample visualization of each clustering.} Each column is from the same cluster.}
    \label{fig:visall}
\end{figure*}

\begin{figure*}
    \centering
    \includegraphics[width=\linewidth]{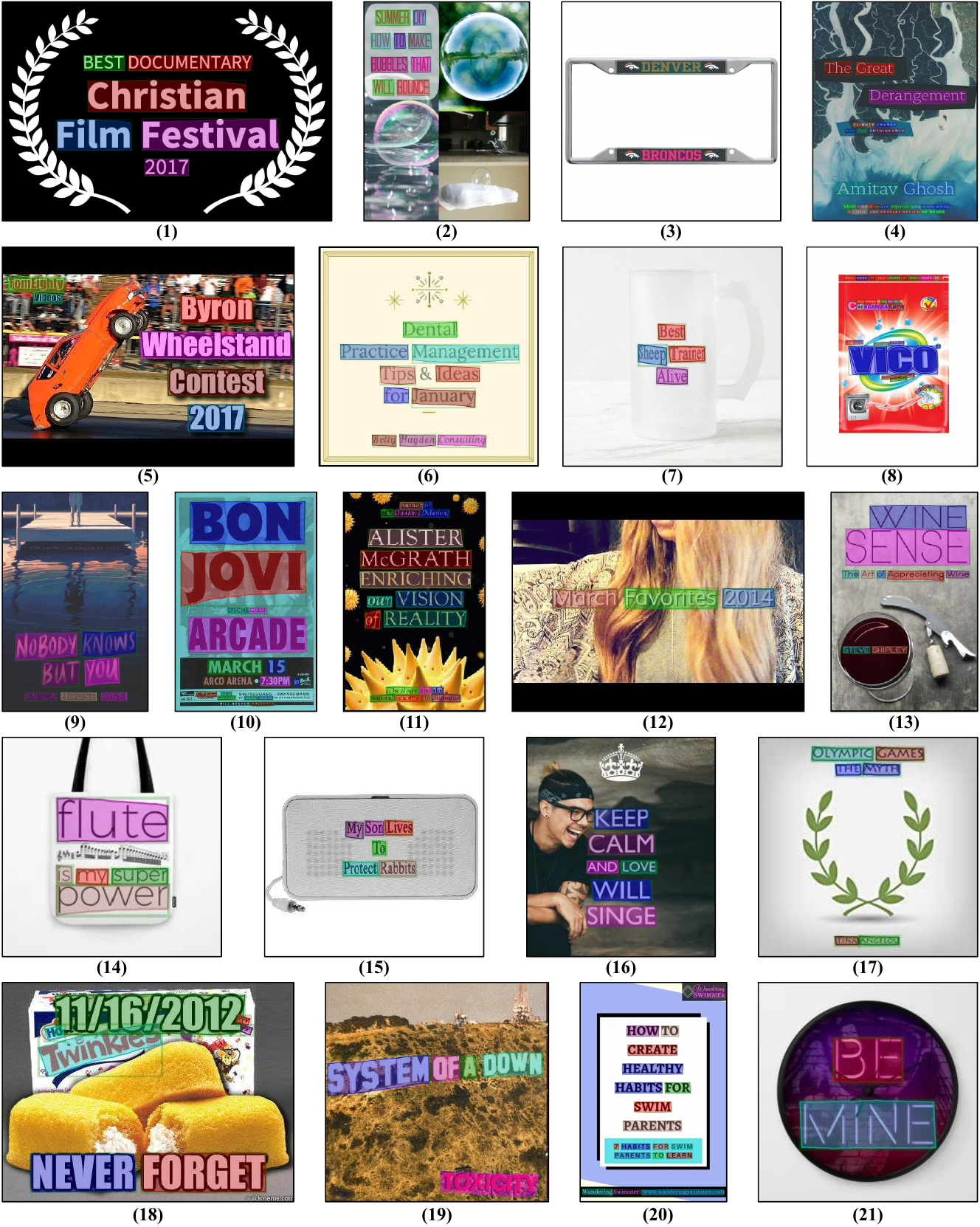}
    \caption{More sample visualization of parrot captions and the corresponding captions are listed in Tab.~\ref{tab:description}}
    \label{fig:vis_samples}
\end{figure*}

In Fig.~\ref{fig:visall}, we visualize samples with top CLIP scores in 250 randomly sampled clusters from the original 4000 clusters.
Each cluster is associated with a certain concept or object.
In Fig.~\ref{fig:vis_samples} and Tab.~\ref{tab:description}, we show more examples with parrot captions and the text spotted results.

\begin{table*}[t]
	\begin{center}
\scalebox{0.9}
{
\begin{tabular}{c|l|l}
            \toprule
            ID & Type & Content \\
            \midrule
            \multirow{2}{*}{1} &
            Captions & BEST DOCUMENTARY - Christian Film Festival - 2017 (1).png\\
            & Co-Emb. & BEST DOCUMENTARY Christian Film Festival 2017\\ 
            \midrule
            \multirow{2}{*}{2} &
            Captions & how-to-make-bubbles-that-bounce\\
            & Co-Emb. & how to make bubbles that bounce\\
            \midrule
            \multirow{2}{*}{3} &
            Captions & Denver Broncos Carbon Small Over Small Metal Acrylic Cut License Plate Frame\\
            & Co-Emb.& Denver Broncos\\        
            \midrule
            \multirow{2}{*}{4} &
            Captions & Title details for The Great Derangement by Amitav Ghosh - Available\\
            & Co-Emb.& The Great Derangement by Amitav Ghosh\\
            \midrule
            \multirow{2}{*}{5} &
            Captions & Byron Wheelstand Contest 2017\\
            & Co-Emb. & Byron Wheelstand Contest 2017\\  
            \midrule
            \multirow{2}{*}{6} &
            Captions & dental marketing and practice management ideas for January - winter dental marketing ideas     betty hayden consulting\\
            & Co-Emb. & dental practice management ideas for January betty hayden consulting\\
            \midrule
            \multirow{2}{*}{7} &
            Captions & Best Sheep Trainer Alive Frosted Glass Mug\\
            & Co-Emb. & Best Sheep Trainer Alive\\
            \midrule
            \multirow{2}{*}{8} &
            Captions & [THQ VIETNAM] VICO AUTOMATIC WASHING POWDER 3KG X 4 PACKS\\
            & Co-Emb. & VICO AUTOMATIC WASHING POWDER\\
            \midrule
            \multirow{2}{*}{9} &
            Captions & Nobody Knows But You by Anica Mrose Rissi\\
            & Co-Emb. & Nobody Knows But You Anica Mrose Rissi\\
            \midrule
            \multirow{2}{*}{10} &
            Captions & Bon Jovi Poster from Arco Arena on 15 Mar 93: 11 x 17 \\
            & Co-Emb. & Bon Jovi  Arco Arena 15\\
            \midrule
            \multirow{2}{*}{11} &
            Captions & Enriching our Vision of Reality de Alister Mcgrath\\
            & Co-Emb. & Alister Mcgrath Enriching our Vision of Reality \\
            \midrule
            \multirow{2}{*}{12} &
            Captions & March Favorites 2014 | FreshExpectations\\
            & Co-Emb. & March Favorites 2014 \\
            \midrule
            \multirow{2}{*}{13} &
            Captions & Wine Sense: The Art of Appreciating Wine by Steve Shipley\\
            & Co-Emb. & Wine Sense: The Art of Appreciating Wine by Steve Shipley\\
            \midrule
            \multirow{2}{*}{14} &
            Captions & Flute is my super power Tote Bag\\
            & Co-Emb. & Flute is my super power\\
            \midrule
            \multirow{2}{*}{15} &
            Captions &My Son Lives To Protect Rabbits Travel Speaker\\
            & Co-Emb. &My Son Lives To Protect Rabbits\\
            \midrule
            \multirow{2}{*}{16} &
            Captions &Poster: KEEP CALM AND LOVE WILL SINGE\\
            & Co-Emb. &KEEP CALM AND LOVE WILL SINGE\\
            \midrule
            \multirow{2}{*}{17} &
            Captions &Olympic Games - The Myth Audiobook by Tina Angelou Narrated by Danae Phelps\\
            & Co-Emb. &Olympic Games The Myth Tina Angelou\\
            \midrule
            \multirow{2}{*}{18} &
            Captions &11/16/2012 never forget - 11/16/2012 never forget  Twinkie RIP\\
            & Co-Emb. &11/16/2012 never forget Twinkie\\
            \midrule
            \multirow{2}{*}{19} &
            Captions &SYSTEM OF A DOWN : TOXICITY [ECO STYLE] (CD)\\
            & Co-Emb. &SYSTEM OF A DOWN TOXICITY\\
            \midrule
            \multirow{2}{*}{20} &
            Captions &Text reads: how to create healthy habits for swim parents. A blue text box below reads: 7 habits for swim parents to learn\\
            & Co-Emb. &how to create healthy habits for swim parents 7 habits for swim parents to learn\\
            \midrule
            \multirow{2}{*}{21} &
            Captions &Be Mine Wall Clock\\
            & Co-Emb. &Be Mine\\
            \bottomrule
            \end{tabular}
            }
	\end{center}
 	\caption{
		Captions and the Co-Emb.~Text shown in Fig~\ref{fig:vis_samples}.
	}
        \label{tab:description}
\end{table*}

\end{document}